\documentclass[12pt,letterpaper]{article}
\usepackage[a4paper, total={7in, 10in}]{geometry}

\usepackage{helvet}
\usepackage{authblk}
\usepackage{hyperref}

\makeatletter
\renewcommand{\maketitle}{\bgroup\setlength{\parindent}{0pt}
\begin{flushleft}
  \textbf{\@title}
  
  \@author
\end{flushleft}\egroup}
\makeatother

\title{Exploiting Noise as a Resource for Computation and Learning in Spiking Neural Networks}
\date{2023.9.1}

\author[1, $\diamondsuit$]{Gehua Ma}
\author[2]{Rui Yan}
\author[1,3 ,$\star$]{Huajin Tang}

\affil[1]{College of Computer Science and Technology, Zhejiang University \& The State Key Lab of Brain-Machine Intelligence, Hangzhou, P.R.C.}

\affil[2]{College of Computer Science and Technology, Zhejiang University of Technology, Hangzhou, P.R.C.}

\affil[3]{Lead Contact}

\affil[$\star$]{  Correspondence: \url{htang@zju.edu.cn} (H.T.)}
\affil[$\diamondsuit$]{ \url{gehuama@{icloud.com, zju.edu.cn} } }

\usepackage[super,comma,sort&compress]{natbib}
\usepackage{amsmath,amssymb}
\usepackage{graphicx}
\usepackage{ntheorem}

\newtheorem{theorem}{Theorem}
\newtheorem{lemma}{Lemma}
\newtheorem*{assumption}{Assumption}

\newtheorem*{proof}{Proof}

\usepackage{txfonts}

\begin{document}

\maketitle

\section*{Summary}

Networks of spiking neurons underpin the extraordinary information-processing capabilities of the brain and have become pillar models in neuromorphic artificial intelligence. Despite extensive research on spiking neural networks (SNNs), most studies are established on deterministic models, overlooking the inherent non-deterministic, noisy nature of neural computations. This study introduces the noisy spiking neural network (NSNN) and the noise-driven learning rule (NDL) by incorporating noisy neuronal dynamics to exploit the computational advantages of noisy neural processing. NSNN provides a theoretical framework that yields scalable, flexible, and reliable computation. We demonstrate that NSNN leads to spiking neural models with competitive performance, improved robustness against challenging perturbations than deterministic SNNs, and better reproducing probabilistic computation in neural coding. This study offers a powerful and easy-to-use tool for machine learning, neuromorphic intelligence practitioners, and computational neuroscience researchers.


\section*{Keywords}

Spiking neural networks, noisy spiking neural networks, surrogate gradients, noise-driven learning, neuromorphic intelligence, neural coding, probabilistic graphical models, dynamical systems

\section*{Introduction}
Spiking neural networks (SNNs) are seen as promising to bridge the gap between artificial and biological neural networks. They have been widely used as computational models in neuroscience research \cite{zilli2010coupled,teeter2018generalized}. Benefiting from the recent progress in deep learning \cite{simonyan2014very,szegedy2015going,he2016deep}, SNNs have also achieved remarkable advantages in various applications like computer vision and robotics \cite{Lee2016,Wu2018,Wu2019,deng2021temporal,volinski2022data,zhao2022nature,fang2021deep,Liu2020,roy2019towards}. In general, the majority of spiking neural models are established by deterministic SNNs (DSNNs), which ignore the inherent randomness of spiking neurons. Spiking neurons with noise-perturbed dynamics are considered more biologically realistic as the ion channel fluctuations and synaptic transmission randomness can cause noisy sub-threshold membrane voltages \cite{verveen1974membrane,kempter1998extracting,stein1965theoretical,stein2005neuronal,faisal2008noise,maass1995noisy,maass1996noisy,patel2005stochastic}. Furthermore, the internal noise incurs a potential benefit in generalization performance by promoting a more fault-tolerant representation space \cite{liu2020does,camuto2020explicit,lim2021noisy} and preventing overfitting \cite{hinton2012improving}. However, a generic and flexible approach is required to fully exploit the noisy spiking neural models and comprehend the role of noise in networks of spiking neurons.

Previous literature \cite{gerstein1964random,tuckwell1989stochastic} has investigated spiking neurons with stochastic activity by including a noise term in the differential equation of the membrane voltage. Such a noise term is typically modeled as white or colored noise derived in the stochastic differential equation with a diffusion process \cite{plesser2000noise,gerstner2014neuronal}. These approaches introduced detailed spiking models with noisy membrane dynamics; however, they do not attempt to build a generic method on the network level. Some studies \cite{rao2004bayesian,rao2004hierarchical,deneve2004bayesian,kasabov2010spike} have presented small networks of noisy spiking neurons that can perform probabilistic inference. Nevertheless, these methods are difficult to incorporate arbitrary network architectures due to the lack of an effective learning method. A recent study \cite{skatchkovsky2021spiking} introduced a Generalized Linear Model variant of the deterministic Spike Response Model, but their method also does not scale to deep SNNs of interest here. In deep SNNs, an ad-hoc solution called Surrogate Gradient Learning \cite{neftci2019surrogate,cramer2022sg,eshraghian2021training} (SGL, pseudo derivative \cite{bellec2020solution}) has become the most widely used to solve the discontinuity when performing backpropagation \cite{rumelhart1986learning}. While the surrogate gradient methods have proven highly effective \cite{zenke2021remarkable}, they lack a theoretical foundation and a rational explanation \cite{jang2019introduction}. In contrast, Neuroscientific-informed learning methods, like STDP \cite{dan2004spike,froemke2005spike,guyonneau2005neurons}, are theoretically grounded and proven promising but struggle to work well in large networks and for complex tasks. It is therefore expected to develop an effective and scalable learning method like SGL while retaining insights regarding learning mechanisms like STDP.

The previous lack of a general computation and learning co-design of noisy spiking neural models has impeded their use. This prevents us from fully exploiting the performance of noisy spiking neural networks as machine learning models and exploring their potential as computational tools for neuroscience. Thus, this article aims to provide a generic and flexible integration of deep SNNs and noisy spiking neural models. In this way, we can make the spiking neural model biologically more realistic and get potential performance gains while being able to directly beneﬁt from the engineering advances emerging in the rapidly-growing deep learning ﬁeld. Moreover, in terms of theoretical value, this also demonstrates how noise may act as a resource for computation and learning in general networks of spiking neurons \cite{maass2014noise}.

Here, we show a noisy spiking neural network model (NSNN) using a noise-driven learning (NDL) paradigm. The approach exposed here provides a generic and flexible integration of noisy spiking neural models and deep SNNs. Additionally, NDL subsumes SGL and provides an insightful explanation for the latter. By incorporating various SNN architectures and algorithms, we demonstrate the effectiveness of NSNNs. Also, NSNNs lead to significantly improved robustness when facing challenging perturbations, such as adversarial attacks. In addition, through NSNN-based neural coding analyses, we show the potential of the NSNN model as a useful computational tool for neuroscientific research.


\section*{Results}

\subsection*{Noisy spiking neural network and noise-driven learning}

\begin{figure}[h]
\centering
    \includegraphics[width=1 \textwidth]{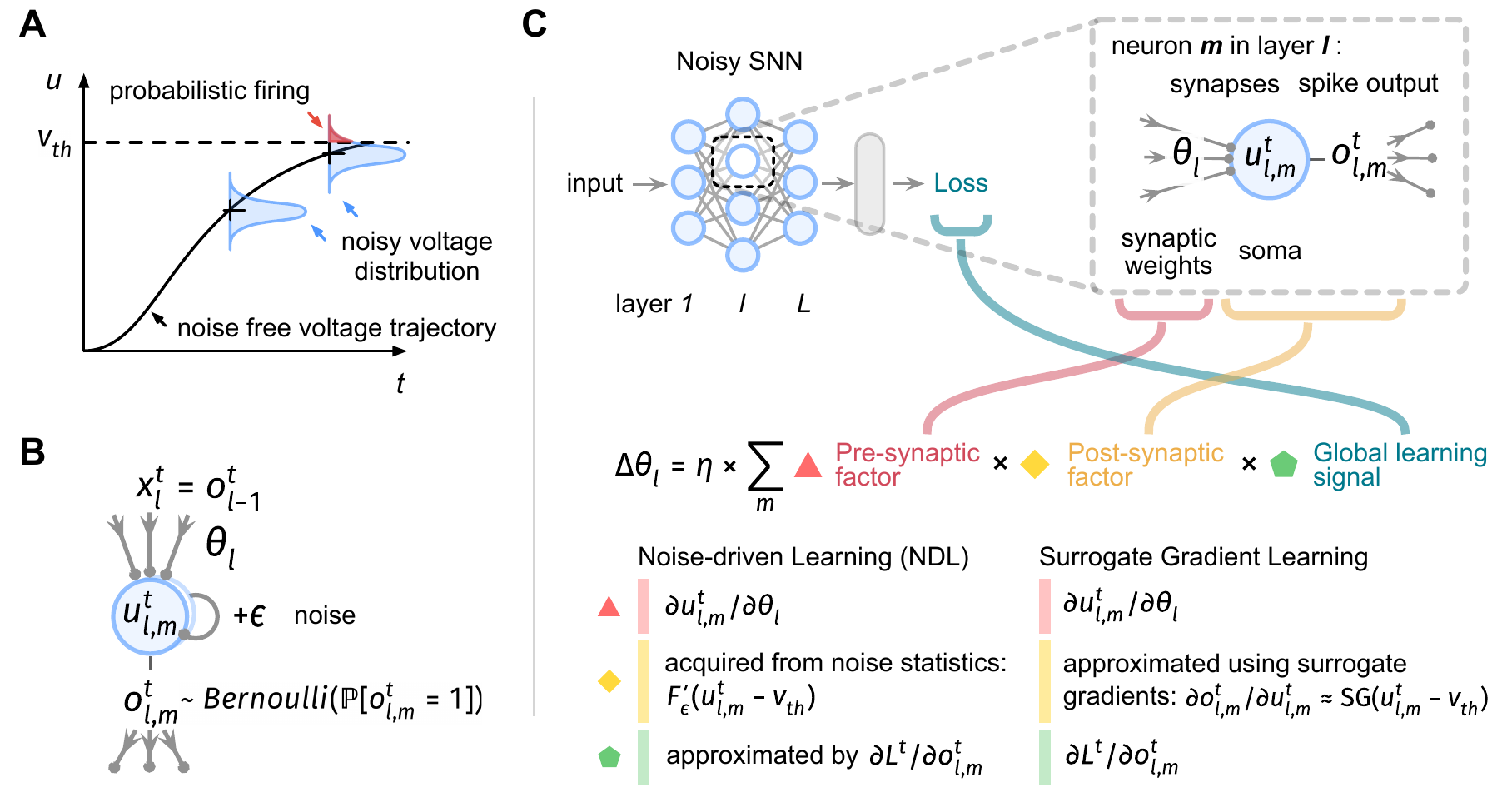}
    \caption{
    	{\bf Schemes for NSNN and NDL. }
	    {\bf A} Introducing noisy neuronal dynamics yields a probabilistic firing mechanism, where the firing probability is given by the membrane noise cumulative distribution function, indicated by the shaded red part under the noisy voltage distribution. Here $u$ denotes membrane voltage and $v_{\text{th}}$ denotes firing threshold. 
    	{\bf B} Computation flow in a Noisy LIF neuron $l,m$ with membrane voltage $u^{t}_{l,m}$, input $\boldsymbol{o}^t_{l-1}$, and output $o^t_{l,m}$. 
	    {\bf C} Illustration of the computation for updating synaptic weights $\theta_l$ of layer $l$ using NDL and SGL. $F_\epsilon$ denotes the cumulative distribution function of noise $\epsilon$, $\eta$ denotes learning rate. 
    }
    \label{fig:1}
\end{figure}

We consider a Noisy Leaky Integrate-and-Fire (LIF) spiking neuron model ({\bf see Experimental Procedures} Method details) in this article, following previous literature that uses diffusive approximation \cite{plesser2000noise,burkitt2006review,gerstner2014neuronal}. It considers a discrete sub-threshold equation of the form:
\begin{equation} \label{eq:nlif-main}
	\text{Noisy LIF sub-threshold dynamic: } u^t = \tau u^{t-1} + \phi_{\theta}(x^t) + \epsilon, 
\end{equation}
where $u$ denotes membrane potential, $x^t$ denotes input at time $t$, $\tau$ is the membrane time constant, and $\phi_\theta$ is a parameterized input transform. The noise $\epsilon$ is drawn from a zero-mean Gaussian distribution. Introducing internal noise induces a probabilistic firing mechanism (illustrated in Figure \ref{fig:1}A), where the difference $u-v_{\text{th}}$ governs the firing probability \cite{maass1995noisy,plesser2000noise,gerstner2014neuronal}: 
\begin{equation} \label{eq:nlif_spike-main}
	\text{Noisy LIF probabilistic firing: } o^t \sim \text{Bernoulli}(\mathbb{P}[o^t=1]),  
\end{equation}
where $o^t$ is the spike state, $\mathbb{P}[o^t=1]=F_\epsilon(u^t - v_{\text{th}})$, $F_\epsilon$ is the cumulative distribution function of the noise, and $v_{\text{th}}$ is the firing threshold. 

By employing Noisy LIF neurons, NSNN presents a general form of spiking neural networks. For instance, if the noise variance $\operatorname{Var}[\epsilon]$ approaches zero, the firing probability function $F_\epsilon$ converges to the Heaviside step function, the Noisy LIF model, therefore, subsumes the deterministic LIF case. This suggests that conventional DSNNs can be viewed as a special type of noisy spiking neural network. Further, by considering logistic membrane noise, the Noisy LIF neuron covers the sigmoidal neuron model \cite{maass2014noise}. 

We then provide a well-defined solution of synaptic optimization via gradient descent for networks of spiking neurons. As noisy neurons code for binary variables, we may represent NSNNs using the Bayesian Network \cite{heckerman1995learning,heckerman1998tutorial}, a probabilistic graphical model representing a set of variables and their conditional dependencies by a directed graph. Leveraging the Bayesian Network formulation, we can succinctly represent the spike states in NSNNs. This enables us to convert the gradient computation during synaptic optimization into gradient estimation in a probabilistic model, thereby circumventing the problematic firing function derivative. The resulting noise-driven learning (NDL) rule ({\bf see Experimental Procedures} Method details) is illustrated in Figure \ref{fig:1}C. In particular, the concise form of NDL makes it easy to combine with other online \cite{bellec2020solution,zenke2021brain} or local learning \cite{wu2022brain} strategies to utilize computational resources more efficiently and handle more diverse tasks.

\begin{figure}[h]
\centering
    \includegraphics[width=1 \textwidth]{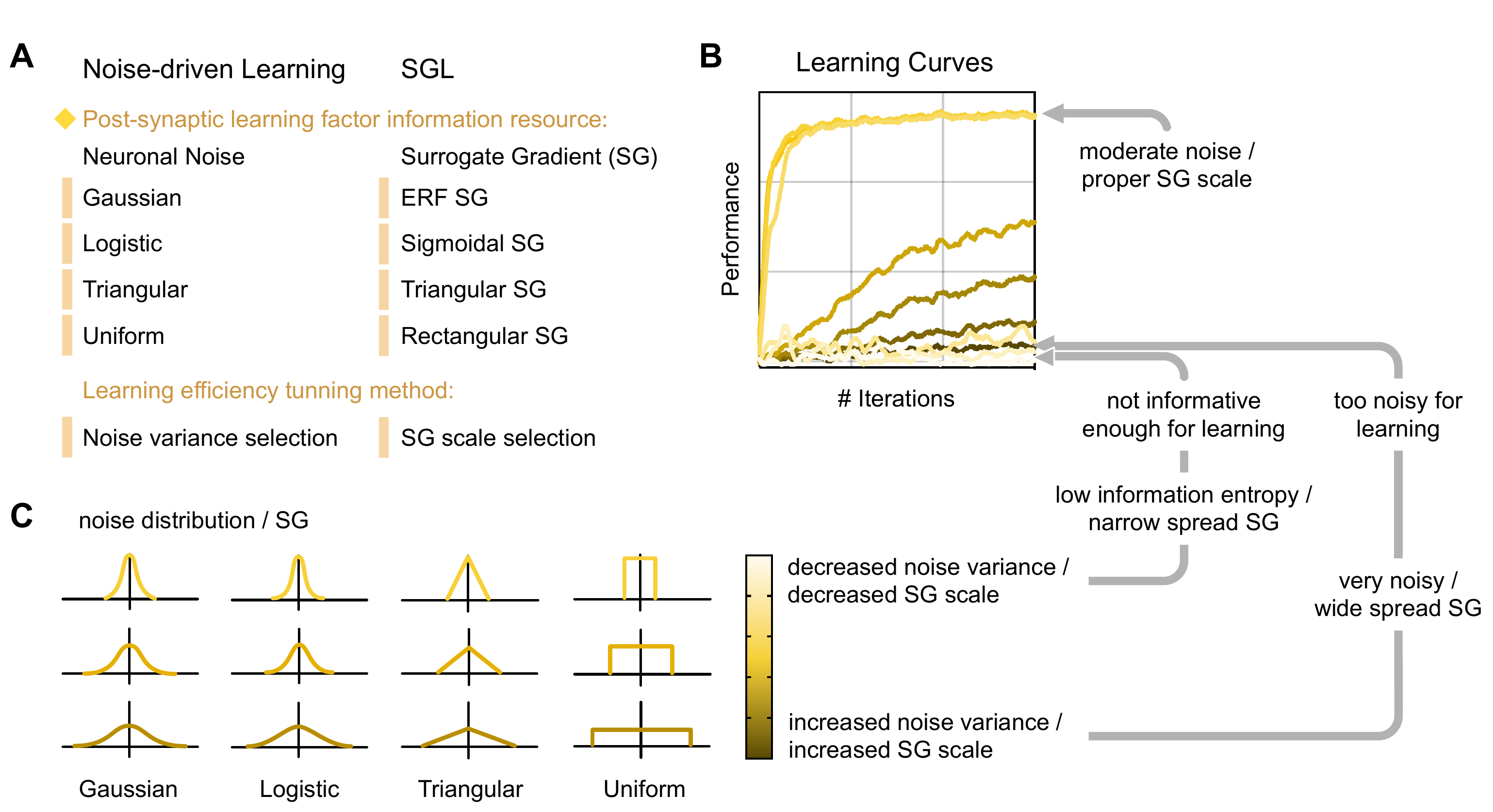}
    \caption{
    	{\bf NDL provides a principled backend for SGL. }
	    {\bf A} The correspondence between various types of neuronal noise and surrogate gradient functions. When using surrogate gradients, we obtain statistical information from the noisy membrane voltage dynamics to form a post-synaptic learning factor.
    	{\bf B, C} The effect of adjusting noise variance or surrogate gradient scale on learning efficiency. 
    }
    \label{fig:2}
\end{figure}

One noteworthy feature of NDL is providing a principled rationale for SGL. By leveraging the three-factor learning rule framework \cite{fremaux2016neuromodulated,gerstner2018eligibility}, we show the mathematical relationship presented in Figure \ref{fig:1}C, indicating that SGL can be regarded as a special type of NDL. Although the surrogate gradients \cite{Wu2018,neftci2019surrogate,zenke2021remarkable,cramer2022sg} have been commonly associated with the straight-through estimator \cite{Hubara2016,tokui2017evaluating,Hou2017,Yin2019} in binarized networks, this association does not justify from the perspective of spiking neurons. In this sense, SGL appears to be an ad-hoc solution rather than a theoretically sound learning method \cite{jang2019introduction,neftci2019surrogate}. NDL reveals the essence of the surrogate gradient, which is to obtain the post-synaptic learning factor from the neuron membrane noise statistics. Figure \ref{fig:2}A illustrates the relationship between noise variance selection in NDL and surrogate gradient scale selection in SGL. When the noise variance is small, the probabilistic inference forward of an NSNN can be approximated by the deterministic forward pass in a DSNN. Therefore, NDL derived within the NSNN framework subsumes SGL in conventional DSNNs. This also provides a random noise explanation for adjusting the scale (shape) of the surrogate gradient functions in SGL. Figures \ref{fig:2}B and \ref{fig:2}C visualize the impact of this variance selection or scale selection process on the learning effect of spiking neural networks, demonstrating that adjusting the scale of the surrogate gradient functions in SGL can be viewed as selecting noises with different variances in NDL.

\subsection*{NSNN leads to high-performance spiking neural models}

To verify the effectiveness and compatibility of NSNNs, we conducted experiments on multiple recognition benchmark datasets using various combinations of SNN architectures and algorithms. Recognition datasets we considered include static image datasets, including CIFAR-10 and CIFAR-100 \cite{krizhevsky2009learning}, and event stream datasets collected using DVS cameras (silicon retina), including DVS-CIFAR \cite{li2017cifar10} and DVS-Gesture \cite{amir2017low}. More experimental details are presented in {\bf Experimental Procedures} Experimental details. 

We analyzed the accuracy of the models on these datasets, and we mainly focus on comparison with SNN models here. As shown in Table \ref{tab:1}, results on CIFAR-10 and CIFAR-100 demonstrate the effectiveness of NSNNs on static image recognition tasks. For instance,  on CIFAR-10, the NSNN model (with STBP, CIFARNet) achieved an accuracy of 0.9390 for two simulation timesteps, while its deterministic counterpart reported 0.9188. Our results indicate that NSNNs work well with various SNN architectures and algorithms, demonstrating great compatibility. Therefore, NSNNs can benefit from more efficient SNN algorithms or architectures. For example, when using the ResNet-18 architecture, the NSNN model using the more efficient TET algorithm performs significantly better than the NSNN model using STBP (Table \ref{tab:1}, CIFAR-10, CIFAR-100). NSNNs also demonstrate effectiveness and compatibility on event stream data recognition tasks (Table \ref{tab:1}, DVS-CIFAR, DVS-Gesture) and outperformed their DSNN counterparts. In particular, on DVS-CIFAR data, the NSNN model with STBP and ResNet-19 significantly outperforms its DSNN counterpart, and performance improvements in other combinations are also evident. Due to the limited number of samples, DSNNs often experience severe overfitting when working with DVS-CIFAR and DVS-Gesture data. However, in NSNNs, internal noise improves the model’s generalization ability, resulting in better performance than their deterministic counterparts. These generalization improvements brought about by internal noise in NSNNs align with previous research results in ANNs \cite{hinton2012improving,camuto2020explicit,lim2021noisy}.

\subsection*{NSNN leads to improved robustness against challenging perturbations}
\label{sec:rob}

Robustness is essential for the information processing of spiking neural models to prevent external perturbations and interferences in real-world environments. From an application standpoint, robustness ensures reliable performance when facing corrupted input (possibly caused by errors in data collection and processing) and perturbed internal information flow (possibly caused by communication abnormalities between different units). In terms of building biologically realistic computational neural circuits, robust spiking neural models align more closely with the noisy yet resilient characteristics of biological circuits \cite{basalyga2006response,faisal2008noise,mcdonnell2011benefits}.

Next, we demonstrate the improvement in robustness achieved by using NSNNs. To this end, we conducted perturbed recognition experiments on CIFAR-10, CIFAR-100, and DVS-CIFAR data ({\bf see Experimental Procedures} Experimental details). We evaluated the performance of DSNNs and NSNNs by measuring their accuracy and loss values under various types and intensities of perturbations. The models were trained as described in the previous recognition experiments and tested using input-level or spike state-level perturbations.  

We considered several challenging input-level perturbations. For static image data CIFAR-10 and CIFAR-100, we used challenging adversarial attacks to construct corrupted inputs. This work leveraged two adversarial attacks: the Fast Gradient Sign Method (FGSM) and the Direct Optimization method (DO). For event stream data DVS-CIFAR, we used the EventDrop \cite{gu2021eventdrop} perturbation, whose basic idea is randomly dropping a proportion of events with a probability of $\rho$. 
We also considered directly perturbing all spike states (firing states of spiking neurons) in SNNs to directly mimic the spike train variability in biological neural circuits. The intensity of the perturbation is controlled by a parameter $\beta$. Details are presented in {\bf Experimental Procedures}. 

\begin{figure}[h]
\centering
    \includegraphics[width=1 \textwidth]{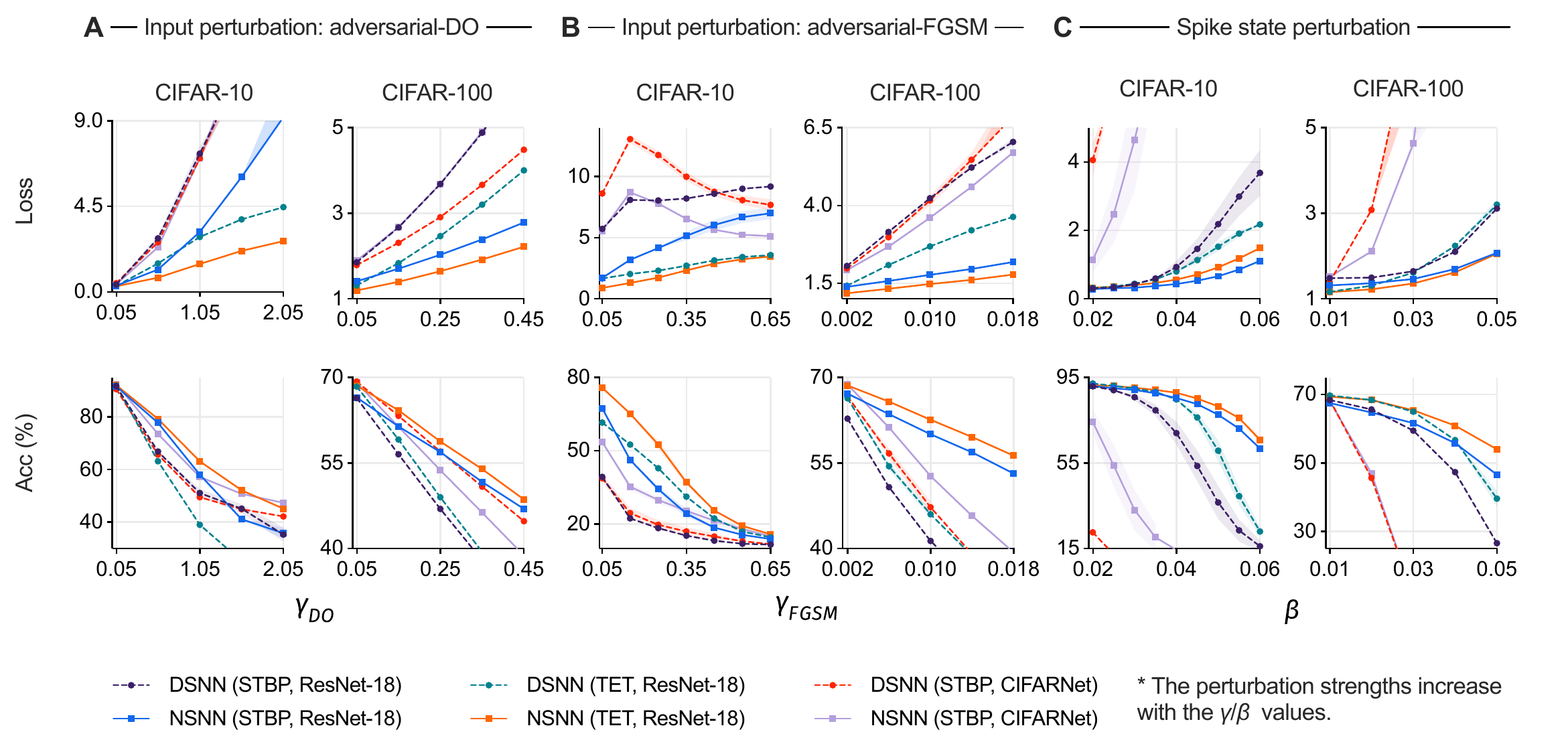}
    \caption{
    	{\bf Results of perturbed recognition experiments CIFAR-10 and CIFAR-100 datasets. }  
    	{\bf A, B} Loss and accuracy results under input-level adversarial attacks (DO method, FGSM method).
    	{\bf C} Loss and accuracy results under spike state-level (firing state of spiking neurons) perturbations.
    }
    \label{fig:3}
\end{figure}

Figure \ref{fig:3} shows that using NSNNs significantly improved robustness for static data recognition tasks (CIFAR-10, CIFAR-100). When facing challenging adversarial attacks and hidden-state perturbations, NSNN models consistently outperformed their DSNN counterparts. For example, in the CIFAR-100 FGSM adversarial attack experiment (Figure \ref{fig:3}B), NSNNs demonstrated good resilience, whereas the reliability of DSNNs degraded radically as the perturbation intensity increased. Similarly, when facing direct interference with neuron spike states (Figure \ref{fig:3}C), NSNNs exhibited superior robustness to their deterministic counterparts.
The perturbed experiments on DVS-CIFAR data also showed that the NSNN model achieved better robustness than DSNNs (Tables \ref{tab:5}, \ref{tab:6}). When facing input-level EventDrop perturbations, NSNNs achieved lower loss and higher accuracy than their deterministic counterparts in most cases (Table \ref{tab:5}). Similarly, as shown in Table \ref{tab:6}, when facing hidden state perturbations, NSNNs demonstrated good resilience. Their superiority becomes apparent as the perturbation (spike state) level increases. For example, at a perturbation level of $\beta=0.01$ (with STBP and ResNet-19), the accuracy of NSNNs was $4.5\%$ higher than that of DSNNs. When the perturbation level increased to 0.04, the accuracy lead of NSNNs over DSNNs reached $46.3\%$. We provide a theoretical analysis of internal noise and model stability in the following text (see also {\bf Experimental Procedures} Theoretical analyses on internal noise and stability). 

\subsection*{NSNN demonstrates a promising tool for neural coding research}
\begin{figure}
\centering
    \includegraphics[width=1 \textwidth]{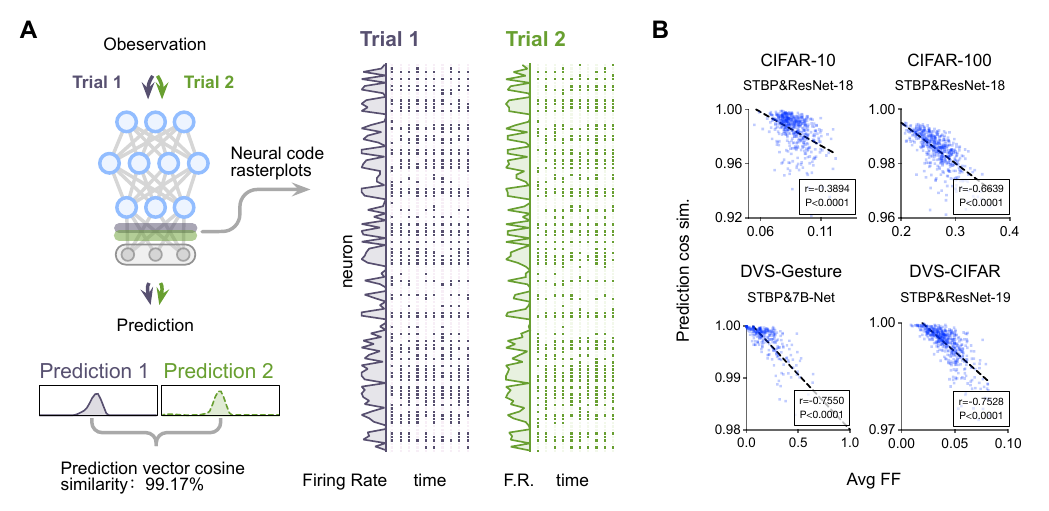}
    \caption{
    	{\bf NSNN-based coding analyses. } 
    	{\bf A} NSNNs exhibit neural code-level variability and decision-level (prediction-level) reliability. We visualize the prediction distributions, spike rates, and raster plots of the final spiking layer outputs of two repeated trials obtained using DSV-CIFAR data.
    	{\bf B} Relationship between the average Fano Factor and prediction cosine similarity. Each dot represents a sample used for computing the Pearson correlation. The dotted line is a linear approximation.
    }
    \label{fig:x1}
\end{figure}

Although capturing the spike-based paradigm in neural circuits, conventional DSNNs failed to account for the reliability and variability in neural spike trains \cite{mainen1995reliability,tiesinga2008regulation}, which limits their application as computational models in neural coding research. By contrast, NSNNs can faithfully recover prediction reliability and neural spike train variability, as shown in Figure \ref{fig:x1}A. The NSNN, therefore, demonstrates a useful tool for investigating neural coding, where to provide information about dynamic sensory cues, the patterns of action potentials in spiking neurons must be variable \cite{de1997reproducibility}. 

We first show how NSNN can be used for neural coding analysis in spiking neural models. In particular, we leverage the NSNN model to provide empirical evidence for a rate coding strategy. Specifically, the internal randomness of NSNNs leads to slightly different neural codes (spike output of the final spiking layer) and predictions between trials given the same input (Figure \ref{fig:x1}A). This allows for analyzing the Pearson correlation between the variation in the firing rate of the neural code and the stability (reliability) of the final prediction in NSNNs. Since we consider a recognition task here, we use the final spiking neuron layer output, which contains more semantically informative data, as the neural code. We use the Fano Factor (FF) to numerically measure the firing rate variation in neural code \cite{edmund2010} and the cosine similarity of prediction vectors to measure prediction stability ({\bf see also Experimental Procedures} Experimental details). A larger FF indicates greater firing rate variation between different trials, while high prediction cosine similarity corresponds to more stable predictive output. We find significant negative correlations between variation in firing rate and stability of prediction in learned NSNNs (Figure \ref{fig:x1}B). These results are robust to various combinations of SNN architectures and algorithms, suggesting that these NSNNs learn a primary rate coding-based policy. This makes sense as membrane noise injection introduces uncertainty into the firing process, reducing the reliability of precise spiking time-based coding. As the same firing rate (represented as firing count in simulation steps here) can correspond to different spike trains, rate-based coding can improve model robustness by constructing a representation space with better fault tolerance.

\begin{figure}
\centering
    \includegraphics[width=1 \textwidth]{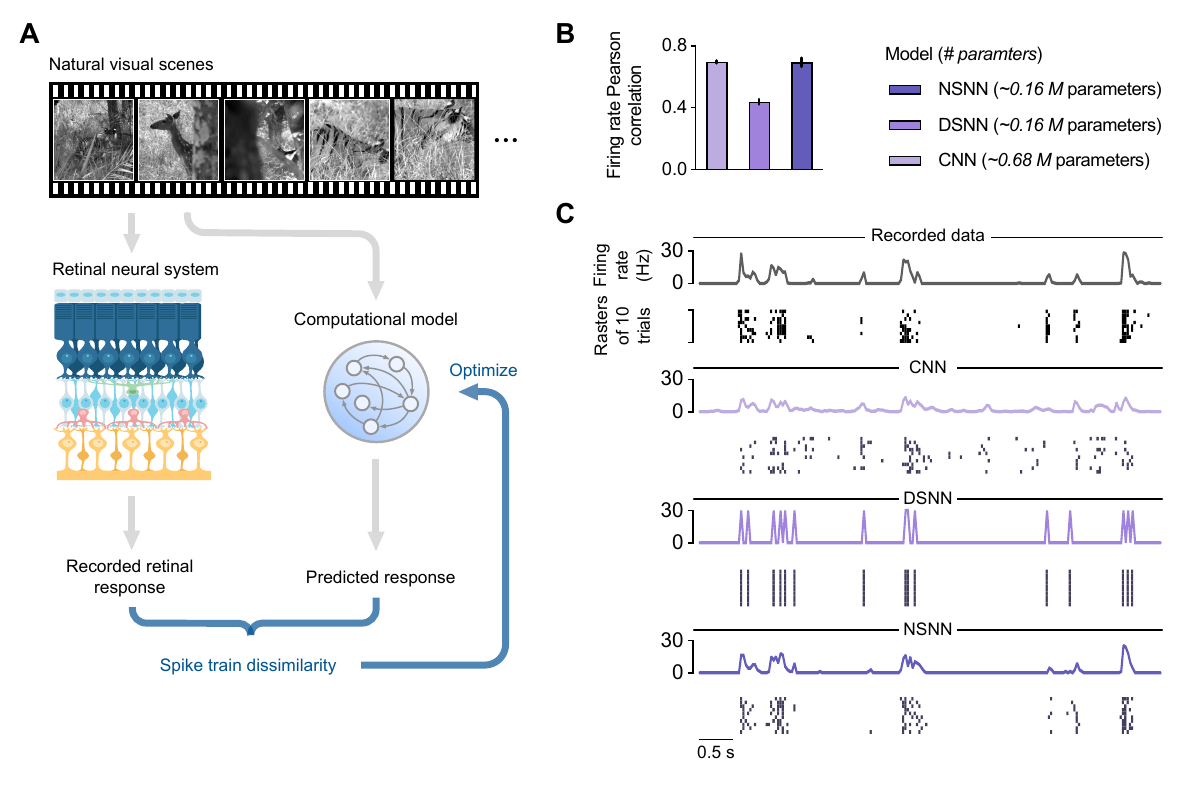}
    \caption{
    	{\bf Neural activity fitting experiments. } 
    	{\bf A} Graphical illustration of the simple neural activity fitting experiment: spiking neural models were optimized to produce spike outputs close to the recorded neural activities.
    	{\bf B} The average firing rate Pearson correlation coefficients of different fitting models. Higher values indicate better predictions. 
    	{\bf C} Spikes and firing rates of the recorded responses of a representative neuron and the predictions of three models (CNN, DSNN, NSNN) to a natural visual scene clip.
    }
    \label{fig:x2}
\end{figure}

Spiking neural models are widely used as computational tools for neural circuits. One effective way to validate these approaches is to simulate the neural processing process on a computer and verify through quantitative testing whether they can perform the tasks completed by some specific biological neural systems and produce similar results. Therefore, we perform a neural activity fitting task ({\bf see also Experimental Procedures} Experimental details) using retinal neural response to natural visual scenes \cite{onken2016using} (Figure \ref{fig:x2}A) to highlight the advantages of NSNN as a computational model. In particular, we constructed DSNN and NSNN models with the same structure for this neural activity fitting task. These spiking neural models took the visual scenes as input and the recorded retinal responses as target outputs and were optimized to produce spike trains similar to recorded neural activity. To compare these spiking neural models with the current state-of-the-art, we also considered a competitive convolutional neural network (CNN) model \cite{mcintosh2016,zheng2021unraveling} as the baseline. We evaluated the Pearson correlation coefficient between the recorded and predicted neural activity to compare the performances of these models numerically. To illustrate that the DSNN model failed to account for variability in the neural activity of interest, we show in Figure \ref{fig:x2}C the spike rasters and firing rates of a representative neuron. As presented in Figure \ref{fig:x2}C, the DSNN model could not reproduce the variable spike patterns and failed to accurately fit real firing rate curves. In contrast, we observed a significant improvement using NSNNs (Figure \ref{fig:x2}B). The NSNN model performs well in fitting both spike activity and firing rate. Additionally, the NSNN model can demonstrate competitive performance with fewer model parameters than the CNN model. These results demonstrate that applying performance optimization to a biologically appropriate NSNN model makes it possible to construct quantitative predictive models of neural coding \cite{yamins2014performance}.

\section*{Discussion}


In this study, we reported noisy spiking neural models and noise-driven learning by exploiting neuronal noise as a resource for computation and learning in networks of spiking neurons. We introduced a membrane noise term into the deterministic spiking neuron model \cite{gerstein1964random,gerstner2002spiking,gerstner2014neuronal}, formulated the networks of these noisy spiking neurons, and derived the NDL learning method to perform synaptic optimization. Our results on multiple datasets indicate that NSNNs exhibit competitive performance and improved generalization ability. Further perturbed tests show that NSNNs demonstrate improved robustness against various perturbations, including challenging adversarial attacks, thereby providing empirical support for the internal noise-model stability analysis. Besides, NSNNs can easily integrate with various SNN algorithms and network architectures, allowing our methodology to be trivially generalized to a wide range of fields that require the presence of stochasticity. Finally, we deliberated from the coding strategy analysis and building predictive coding models as an illustration and demonstrated that NSNNs provide a promising tool for neural coding research. Furthermore, since the general framework presented in this study subsumes traditional deterministic spiking models, our method is expected to enable more flexible and robust computation on neuromorphic hardware with inherent unreliabilities.

\subsection*{Advancing spiking neural network research}
This work presents a novel approach for constructing and training spiking neural networks. The NSNN framework enables the flexible integration of internal noise with different distributions and variances into SNN models. In machine learning, introducing internal noise typically offers performance benefits, such as preventing overfitting and enhancing model robustness. As such, NSNNs can be utilized to obtain better performance models. The computation and learning form of NSNN incorporated with NDL is concise and straightforward, making it easy to implement in larger network architectures using simple module replacement based on the DSNN implementation. Here we demonstrate that the combination of NSNN and NDL can be well incorporated with representative network architectures and algorithms. Going forward, NSNNs can easily be integrated with larger models, like spiking transformers \cite{vaswani2017attention,zhang2022spiking}. Although we currently assign credits along the temporal dimension using backpropagation through time in experiments, NDL is compatible with potential online or local learning methods. Further research into the noisy spiking neural models may provide insights for designing local, online learning mechanisms \cite{bellec2020solution,zenke2021brain,wu2022brain} and combining them with NDL to form more biophysically-plausible SNN learning methods. As such, NSNN presents a promising avenue for advancing spiking neural networks and their learning methods. 

\subsection*{Incorporation with neuromorphic hardware}
As a crucial component of neuromorphic intelligence research \cite{roy2019towards}, SNNs are expected to achieve recognition, adaptation, behavior, and learning at low power consumption through integration with neuromorphic hardware \cite{indiveri2015memory,pei2019towards,davies2021advancing}. The NSNN and NDL approach introduced in this article will benefit hardware implementation of spiking neural models on analog, digital, or mixed \cite{benjamin2014neurogrid} platforms. Neuromorphic hardware such as TrueNorth \cite{debole2019truenorth} and Intel Loihi \cite{loihi2018} incorporate built-in pseudorandom number generators or random noise parameters to simulate the unreliability in computation. This enables the implementation of noisy spiking neural models on neuromorphic hardware and extends the computational power of noisy spiking neural models by utilizing these built-in random sources for practical applications. Furthermore, due to the adoption of noisy neuronal dynamics, NSNNs have considerable robustness (demonstrated by previous experiments) against internal disturbances. Therefore, NSNNs are expected to exhibit more stable performance in neuromorphic hardware systems that inherently have noise \cite{qiao2015reconfigurable,hazan2022neuromorphic}, such as electronic disturbances, input perturbations, and communication errors. This allows for more efficient and effective employment of various neuromorphic hardware developments in real-world scenarios. Hence, this work contributes a fundamental theoretical foundation for the future development of neuromorphic computing.

\subsection*{Noise in biological neural networks}
Noise in biological neural networks originates from diverse sources, including voltage or ligand-gated ion channels, the fusion of synaptic vesicles, and the diffusion of signaling molecules to receptors \cite{faisal2008noise}. Channel noise can also affect membrane potential, spike initiation, and spike propagation in small axons and cell bodies \cite{white2000channel}. Additionally, synapses exhibit randomness in the number of transmitter molecules released in a vesicle and in the diffusion processes of molecules \cite{faisal2008noise}. Early neuroscientists recognized that intrinsic noise in brain activity could confer benefits by randomizing neuronal dynamics, leading to advantages in creativity, probabilistic decision-making, unpredictability, and allocation to discrete categories \cite{edmund2010}. Modern brain signal acquisition and imaging techniques such as EEG, iEEG, and fMRI allow us to explore the role of noise as a computing element in brain networks by pinpointing the locations of different cognitive processes. For instance, a recent study \cite{grossman2019noisy} used iEEG measurement to demonstrate that resting-state cortical noise may influence visual recognition ability. This suggests that noise is not merely a concomitant of neural processing \cite{edmund2010}, but can play a unique and important role in brain networks. By utilizing the NSNN framework in conjunction with brain imaging and electrophysiological signal acquisition techniques, we can establish a tractable computational link between neuronal activity variability and probabilistic behavior. This could further improve our understanding of the role of noise in neural processing and contribute to the comprehension of the mechanisms underlying memory and decision-making in the brain.

\subsection*{Implications to neuroscientific research} 
Spiking neural models are popular in neuroscience research for their biologically realistic spike-based computation paradigm. However, conventional deterministic spiking neural models cannot account for the variability in neural spike trains \cite{de1997reproducibility}. Recent neuroscience research \cite{grossman2019noisy} suggests resting-state cortical noise as a possible neurophysiological trait that limits recognition capacity, implying that our brain is non-deterministic. To understand and simulate cognitive functions such as memory, recognition, attention, and decision in the noisy brain, we are dealing with large-scale, sophisticated computational systems. Modeling these systems requires more complex neural networks composed of a large number of neurons. Therefore, research into the scalable noisy spiking neural model, which aims to provide a useful tool at a computational level, has practical benefits for computational neuroscience. NSNNs are expected to be employed for neuron type \cite{masland2004neuronal,teeter2018generalized}, neural system identification \cite{klindt2017neural,zhuang2021unsupervised}, and constructing predictive counterparts of neural circuits \cite{cadena2019deep,ratan2021computational}. In experiments exposed in this article, we mainly focus on the potential application of NSNNs in neural coding. This a popular research topic in neuroscience and other fields such as computer vision, neuromorphic computing, neural prosthesis, and brain-computer interface. Our results indicate that NSNNs are able to recover reliability and variability in neural circuits. Notably, NSNNs achieved competitive performance with fewer parameters than conventional ANN models in fitting neural responses to natural visual scenes. As such, NSNN provides a promising tool for building computational accounts for various sensory neural circuits and will enable richer models of complex neural computations in the brain. 


\section*{EXPERIMENTAL PROCEDURES}

\subsection*{Resource availability}

\subsubsection*{Lead contact}

Further information and requests for resources should be directed to and will be fulfilled by the Lead Contact, Prof. Huajin Tang (\url{htang@zju.edu.cn}).

\subsubsection*{Materials availability}

This study did not generate new unique reagents.

\subsubsection*{Data and code availability}
The data used in this paper are publicly available and can be accessed at \url{https://www.cs.toronto.edu/~kriz/cifar.html} for the CIFAR (CIFAR-10, CIFAR-100) dataset, \url{https://research.ibm.com/interactive/dvsgesture/} for the DVS128 Gesture (DVS-Gesture) dataset, and \url{https://figshare.com/articles/dataset/CIFAR10-DVS_New/} for the CIFAR10-DVS (DVS-CIFAR) dataset. 

The codebase \cite{ma_2023_7986394} can be found at \url{github.com/genema/Noisy-Spiking-Neuron-Nets} or \url{gitee.com/ghma/Noisy-Spiking-Neuron-Nets} .
Any additional information required to reanalyze the data reported in this paper is available from the lead contact upon request.

\subsection*{Method details}
\subsubsection*{Notations}
We use $x, u, o$ to represent input, membrane potential, and spike output. Also, $x_{l, m}^t, u_{l, m}^t, o_{l,m}^t$ denote variables of neuron $m$ in layer $l$ (whose dimension is $\dim(l)=\dim(\boldsymbol{x}_l)$) at time $t$, where $l \in [1, L]$ and $t \in [1, T]$. We also use boldface type to represent the sets or matrices of variables, e.g., variables of layer $l$ at timestep $t$ are marked as $\boldsymbol{x}_l^t, \boldsymbol{u}_l^t, \boldsymbol{o}_l^t$. Spike state space is marked as $\mathbb{S}$. Notations $\mathbb{E}[\cdot]$, $\mathbb{P}[\cdot]$, $p(\cdot)$ and $F(\cdot)$ are, respectively, expectation, probability, probability distribution, and cumulative distribution function. 

\subsubsection*{LIF neuron}
We consider the commonly used Leaky Integrate-and-Fire (LIF) neuron model \cite{tal1997computing,brunel2007lapicque} in this work, which describes the sub-threshold membrane potential dynamics as 
\begin{equation}
\begin{aligned}
	\tau_m \frac{\mathrm{d}u}{\mathrm{d}t} = -(u-u_{\text{reset}}) + R I(t), u < v_{\text{th}},
\end{aligned}
\end{equation}
where $R,\tau_m$ are membrane resistance and time constant, $I$ is the input current and $v_{\text{th}}, u_{\text{reset}}$ are firing threshold, resting potential, respectively. It leads to the following discrete-time computational form in practice \cite{Wu2018,xiao2022online}:
\begin{equation} \label{eq:lif}
\begin{aligned}
	&\text{sub-threshold dynamic: } u^t = \tau u^{t-1} + \phi_{\theta}(x^t), 	
\\
	&\text{firing: } o^t = \text{spike}(u^t, v_{\text{th}}) \triangleq \text{Heaviside}(u^t - v_{\text{th}}), 	
\\
	&\text{resetting: } u^t = u_{\text{reset}} \text{ if } o^t = 1 , 
\end{aligned}
\end{equation}
where $x^t$ is the input at time $t$, $\tau$ is the membrane time constant, $\phi_\theta:\mathbb{S}^{\dim(x^t)} \rightarrow \mathbb{R}$ denotes a parameterized input transform. To introduce a simple model of neuronal spiking and refractoriness, we assume $v_{\text{th}}=1$, $\tau=0.5$ and $u_{\text{reset}}=0$ throughout this research.

\subsubsection*{Noisy LIF neuron}

The Noisy LIF presented here is based on previous works that use diffusive approximation \cite{plesser2000noise,burkitt2006review,gerstner2014neuronal}, where the sub-threshold dynamic is described by the Ornstein-Uhlenbeck process \cite{van1992stochastic,kloeden1992stochastic}: 
\begin{equation} \label{eq:nlif_ct}
\begin{aligned}
	&\tau_m \frac{\mathrm{d}u}{\mathrm{d}t} = -(u-u_{\text{reset}}) + R I(t) + \xi(t),
	\text{  e.q. } 
	\\
	&\mathrm{d}u = -(u-u_{\text{reset}})\frac{\mathrm{d}t}{\tau_m} + R I(t) \frac{\mathrm{d}t}{\tau_m} + \sigma \mathrm{d}W^t,
\end{aligned}
\end{equation}
the white noise $\xi$ is a stochastic process, $\sigma$ is the amplitude of the noise and $\mathrm{d}W^t$ are the increments of the Wiener process in $\mathrm{d}t$ \cite{gerstner2014neuronal}. As $\sigma\mathrm{d}W^t$ are random variables drawn from a zero-mean Gaussian, this formulation is directly applicable to discrete-time simulations. Specifically, using the Euler-Maruyama method, we get a Gaussian noise term added on the right-hand side of Equation \ref{eq:lif}. Without loss of generality, we extend the additive noise term in the discrete form to general continuous noise \cite{barndorff2001non}, by Equation \ref{eq:lif}, the sub-threshold dynamic of Noisy LIF can be described as:
\begin{equation} \label{eq:nlif}
	\text{Noisy LIF sub-threshold dynamic: } u^t = \tau u^{t-1} + \phi_{\theta}(x^t) + \epsilon, 
\end{equation}
where the noise $\epsilon$ is independently drawn from a known distribution that satisfies $\mathbb{E}[\epsilon] = 0$ and  $p(\epsilon) = p(-\epsilon)$. Equation \ref{eq:nlif} can also be obtained by discretizing an Itô stochastic differential equation variant of LIF neurons \cite{patel2005stochastic,patel2008stochastic}. We consider in this article Gaussian noise for all Noisy LIF neurons in this study. 

The membrane potentials and spike outputs become random variables due to the injection of random noises. Using noise as a medium, we naturally obtain the firing probability distribution of Noisy LIF based on the threshold firing mechanism:
\begin{equation} 
\begin{aligned}
	&\mathbb{P}[\text{firing at time } t] 
	= \mathbb{P}\underbrace{[u^t+\epsilon > v_{\text{th}}]}_{\text{Threshold firing}} 
	\\
	= &\underbrace{\mathbb{P}[\epsilon < u^t-v_{\text{th}}] \triangleq F_\epsilon(u^t-v_{\text{th}})}_{\text{Cumulative Distribution Function definition}}.
\end{aligned}
\end{equation}
Therefore, 
\begin{eqnarray} \label{eq:ot}
o^t=
\begin{cases}
	1, \text{with probability } F_\epsilon(u^t - v_{\text{th}}),
	\\
	0, \text{with probability } \left(1 - F_\epsilon(u^t - v_{\text{th}}) \right).
\end{cases}
\end{eqnarray}
The expressions above show how noise acts as a resource for computation \cite{maass2014noise}. Thereby, we can formulate the firing process of Noisy LIF as
\begin{equation} \label{eq:nlif_spike}
	\text{Noisy LIF probabilistic firing: } o^t \sim \text{Bernoulli} \big( F_\epsilon(u^t - v_{\text{th}}) \big). 
\end{equation}
Specifically, it relates to previous literature on escape noise models \cite{plesser2000noise,plesser2000escape2,jolivet2006predicting}, in which the difference $u-v_{\text{th}}$ governs the neuron firing probabilities \cite{maass1995noisy,gerstner2014neuronal}. In addition, Noisy LIF employs the same resetting mechanism as the LIF model. Unless otherwise indicated, we focus on the discrete form in this research, which is of practical interest.

\subsubsection*{Noisy Spiking Neural Network}

\begin{figure}
\centering
    \includegraphics[width=1 \textwidth]{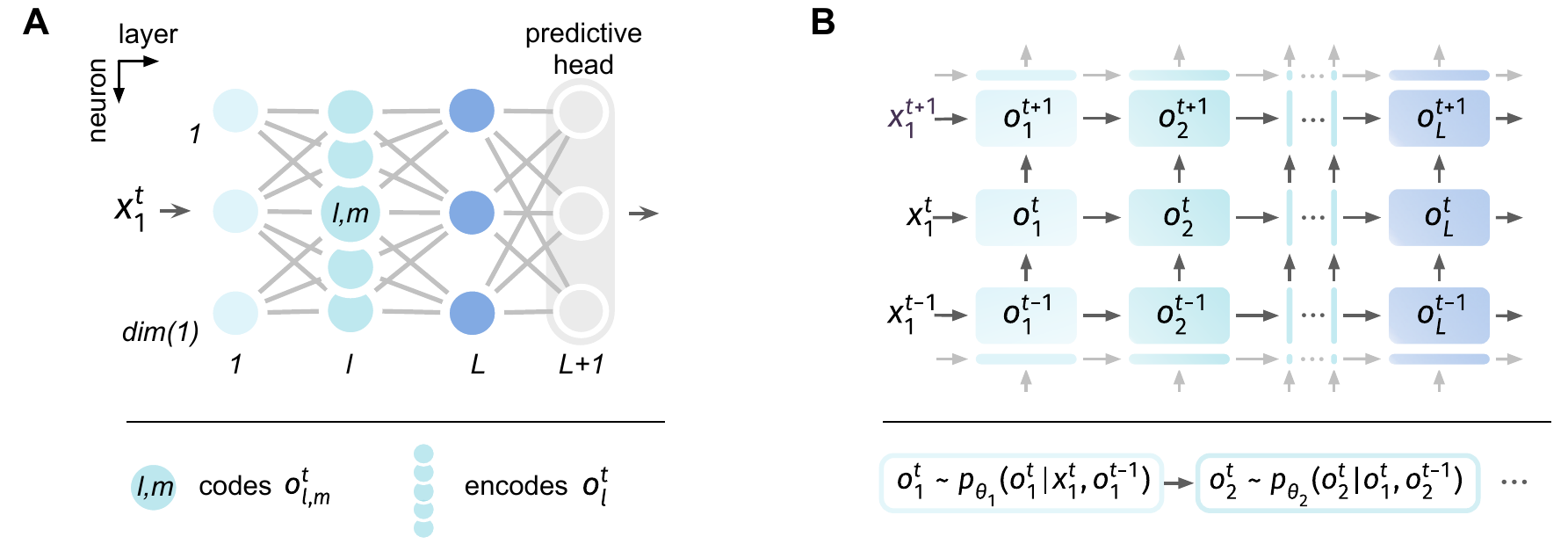}
    \caption{
    	{\bf Graphical illustrations of NSNN. }
    	{\bf A} NSNN graphical illustration with neuron, layer notations, where $x_1$ denotes the initial input to the network.
    	{\bf B} Assumed dependencies between spike states using the Bayesian Network frame. Take time $t$ as an example, the joint distribution is given by $p_{\theta} (\boldsymbol{o}^t_{1 \dots L} | x_1^t, \boldsymbol{o}^{t-1}_{1 \dots L})
	= p_{\theta_1} (\boldsymbol{o}_1^t | x_1^t, \boldsymbol{o}_1^{t-1}) 
	\prod_{l=2}^L p_{\theta_l} (\boldsymbol{o}_l^t | \boldsymbol{o}_{l-1}^t, \boldsymbol{o}_l^{t-1})$, where $\theta$ denotes the synaptic parameters of the parametric model. 
    }
    \label{fig:nsnnbayes}
\end{figure}

We consider the NSNN model as a probabilistic recognition model here as an example (Figure \ref{fig:nsnnbayes}A). Let $x^t_1$ denote the input at time $t$, using the dynamics of Noisy LIF in Equations \ref{eq:nlif}-\ref{eq:nlif_spike}, an NSNN consists of $L+1$ layers is given by
\begin{equation} \label{eq:nsnn} 
\begin{aligned} 
	& \text{input layer: } 
	\boldsymbol{x}^t_1 = x^t_1, 
	\boldsymbol{u}^t_1 = \tau \boldsymbol{u}^{t-1}_1 + \Phi_{\theta_1}(\boldsymbol{x}^t_1) + \boldsymbol{\epsilon}_1,  
	\boldsymbol{o}^t_1 = \big\{o_{1,m}^t: o_{1,m}^t \sim \text{Bernoulli}(\mathbb{P}[o_{1,m}^t = 1]) \big\}_{m=1}^{\dim(1)},
\\
	& \text{hidden layer: } 
	\boldsymbol{x}^t_l = \boldsymbol{o}_{l-1}^t, 
	\boldsymbol{u}^t_l = \tau \boldsymbol{u}^{t-1}_l + \Phi_{\theta_l}(\boldsymbol{x}^t_l) + \boldsymbol{\epsilon}_l, 
	\boldsymbol{o}^t_l = \big\{o_{l,m}^t: o_{l,m}^t \sim \text{Bernoulli}(\mathbb{P}[o_{l,m}^t = 1]) \big\}_{m=1}^{\dim(l)},
\\
	& \text{predictive head: }
	\mathcal{L} = f_{\theta_{L+1}}(\boldsymbol{o}^t_L) = f(\Phi_{\theta_{L+1}}(\boldsymbol{o}^t_L)).
\end{aligned}
\end{equation}
The output $\boldsymbol{o}^t_l$ of layer $l$ is a representation vector in the spike space $\mathbb{S}^{\dim(l)}$, the membrane potentials $\boldsymbol{u}^t_l \in \mathbb{R}^{\dim(l)}$, and the mapping $\Phi_{\theta_l}: \mathbb{S}^{\dim(l-1)} \rightarrow \mathbb{R}^{\dim(l)}$. The noise vector $\boldsymbol{\epsilon}_l \in \mathbb{R}^{\dim(l)}$ consists of independent random noise with a known distribution (Gaussian in this article). The predictive head $f_{\theta_{L+1}}(\boldsymbol{o}^t_L)$ includes a parameterized mapping $\phi_{\theta_{L+1}}(\boldsymbol{o}^t_L)$ and a loss function $f$, denoting the part that decodes predictions from the neural representation $\boldsymbol{o}^t_L$ and compute the loss value. $\phi_{\theta_l}$ represents a map like fully connected or convolution and is thus differentiable w.r.t. parameter $\theta_l$. Dividing the synaptic parameters by layers, as mentioned above, results in no loss of generality as they can be defined as any differentiable mapping. 

For example, to solve recognition problems, we shall consider the predictive probability model $p_{\theta_{L+1}}(y|\boldsymbol{o}^t_L) = \text{softmax}(\phi_{\theta_{L+1}} (\boldsymbol{o}^t_L))$, where the map $\phi_{\theta_{L+1}}$ computes the predictive scores using the neural representation $\boldsymbol{o}^t_L$. The function $f$ can be the cross-entropy of the predictive distribution $p_{\theta_{L+1}}(y|\boldsymbol{o}^t_L)$ and the target distribution $p_{\text{target}}(y|x_1^t)$.  Note that $f_{\theta_{L+1}} (\boldsymbol{o}^t_L)$ here computes the instantaneous loss, different from the $\frac{1}{T}\sum_t f^t$, which is computed over the entire time window and ignores potential online learning \cite{xiao2022online}. 

Since each neuron codes for a random variable $o_{l,m}^t$, we can describe the NSNN by the Bayesian Network model (Figure \ref{fig:nsnnbayes}B) and represent the joint distribution of all spike states given input $x^t_1$ as 
\begin{equation} \label{eq:nsnn_bayesnet}
\begin{aligned} 
	p_{\theta} (\boldsymbol{o}^t_{1 \dots L} | x_1^t, \boldsymbol{o}^{t-1}_{1 \dots L})
	= p_{\theta_1} (\boldsymbol{o}_1^t | x_1^t, \boldsymbol{o}_1^{t-1}) 
	\prod_{l=2}^L p_{\theta_l} (\boldsymbol{o}_l^t | \boldsymbol{o}_{l-1}^t, \boldsymbol{o}_l^{t-1}),
\end{aligned}
\end{equation}
where the layer representation is $p_{\theta_l} (\boldsymbol{o}_l^t | \boldsymbol{o}_{l-1}^t, \boldsymbol{o}_l^{t-1}) = \prod_{m=1}^{\dim(l)} p_{\theta_l} (o_{l,m}^t | \boldsymbol{o}_{l-1}^t, o_{l,m}^{t-1})$.

\subsubsection*{Noise-driven Learning}

To perform NSNN synaptic optimization, the central problem is to estimate the gradient of the expected loss $\mathbb{E}_{\boldsymbol{o}^t_{1 \cdots L}}[\mathcal{L}]$. By Equations \ref{eq:nsnn}, \ref{eq:nsnn_bayesnet}, $g_l$ is given by:
\begin{equation} \label{general_g_l}
	g_l 
	= \nabla_{\theta_l} \mathbb{E}_{\boldsymbol{o}^t_{1 \cdots L}}[\mathcal{L}]
	=
	\nabla_{\theta_l} \sum_{\boldsymbol{o}^t_{1\dots L}} p_\theta(\boldsymbol{o}^t_{1\dots L} | x_1^t, \boldsymbol{o}_{1\dots L}^{t-1}) f_{\theta_{L+1}} (\boldsymbol{o}^t_L).
\end{equation}
As Equation \ref{general_g_l} is intractable to compute, we expect an estimation so that the parameters can be tuned using gradient-based routines. 

The dimensionality of the spike state space is rather limited (either spike or silence). Leveraging this property, we can derive an estimator by conditioning (local marginalization), which performs exact summation over a single random variable to reduce variance \cite{burt1971conditional,aueb2015local}. We first factorize the joint distribution $p_\theta(\boldsymbol{o}^t_{1\cdots L} | x^t_1, \boldsymbol{o}^{t-1}_{1\cdots L})$ as the product of $\prod_{i\neq l} p_{\theta_i} (\boldsymbol{o}^t_{i} | \boldsymbol{o}^t_{i-1}, \boldsymbol{o}^{t-1}_{i})$, $\prod_{k \neq m} p_{\theta_l} (o_{l,k}^t | \boldsymbol{o}^t_{l-1}, o^{t-1}_{l,k})$ and $p_{\theta_l}(o^t_{l,m}|\boldsymbol{o}^t_{l-1},o^{t-1}_{l,m}) $. Then, Equation \ref{general_g_l} becomes
\begin{equation} \label{general_gl2} 
\begin{aligned}
	g_l = \sum_{\boldsymbol{o}^t_{1 \dots L}} \sum_{m} 
	\Big( \prod_{i\neq l} p_{\theta_i} (\boldsymbol{o}^t_i | \boldsymbol{o}^t_{i-1}, \boldsymbol{o}^{t-1}_i)
	\prod_{k \neq m} p_{\theta_l} (o^t_{l,k}|\boldsymbol{o}^t_{l-1}, o^{t-1}_{l,k}) \Big)
	\nabla_{\theta_l} p_{\theta_l} (o_{l,m}^t | \boldsymbol{o}_{l-1}^t, o_{l,m}^{t-1}) f_{\theta_{L+1}} (\boldsymbol{o}^t_L). 
\end{aligned}
\end{equation}
Using the fact $\mathbb{P}[o^t_{l,m}=0] = 1-\mathbb{P}[o^t_{l,m}=1]$, we have that
\begin{equation} \label{general_cond}
\begin{aligned}
	\sum_{o_{l,m}^t} \nabla_{\theta_l} p_{\theta_l} (o_{l,m}^t | \boldsymbol{o}_{l-1}^t, o_{l,m}^{t-1}) f_{\theta_{L+1}} (\boldsymbol{o}^t_L) 
	= \nabla_{\theta_l} p_{\theta_l} (o_{l,m}^t | \boldsymbol{o}_{l-1}^t, o_{l,m}^{t-1}) 
	\Delta\mathcal{L},
\end{aligned}
\end{equation}
where the loss difference term $\Delta \mathcal{L}=f_{\theta_{L+1}}(\boldsymbol{o}^t_L) - f_{\theta_{L+1}} ( \boldsymbol{o}_{\bar{l,m}}^t )$, here we use $\boldsymbol{o}^t_{\bar{l,m}}$ to denote the new state $\boldsymbol{o}^t_L$ if $o_{l,m}^t$ changes. Given that $\sum_{o_{l,m}^t} p_{\theta_l} (o_{l, m}^t) = 1$ and using Equations \ref{general_gl2}, \ref{general_cond}, we have 
\begin{equation}
	g_l = \sum_{\boldsymbol{o}^t_{1 \dots L}} \Big( \prod_{i=1}^L p_{\theta_i} (\boldsymbol{o}_i^t | \boldsymbol{o}_{i-1}^t, \boldsymbol{o}_i^{t-1}) \Big) \hat{g}_l
	= \mathbb{E}_{\boldsymbol{o}^t_{1\dots L}} \left[ \hat{g}_l \right], 
\end{equation}
where
\begin{equation} \label{general_estimator}
\begin{aligned} 
	\hat{g}_l = \sum_m {\nabla_{\theta_l} p_{\theta_l} (o_{l,m}^t | \boldsymbol{o}_{l-1}^t, o_{l,m}^{t-1})} {\Delta\mathcal{L} }. 
\end{aligned}
\end{equation}
Intuitively, this local gradient is defined as a sum of the contributions of all neurons in layer $l$. To get an estimate of $g_l$, we can simply sample from $p_{\theta}(\boldsymbol{o}^t_{1\cdots L} | \boldsymbol{x}^t)$ and calculate using Equation \ref{general_estimator}. However, it is still unwise to compute $\Delta\mathcal{L}$ as it requires repeated evaluations and, thus, cannot scale to large models \cite{tokui2017evaluating}. Inspired by previous studies \cite{fiete2006gradient,tokui2017evaluating,shekhovtsov2020path}, we may attribute the change of loss value to the state flip of variable $o_{l,m}^t$. By doing so, we may approximate the loss difference $\Delta\mathcal{L}$ when the state of $o_{l,m}^t$ alters using a first-order approximation:
\begin{equation} \label{L_approx} 
	\begin{aligned}
	{\Delta\mathcal{L} } \approx 
	\big( o_{l,m}^t - (1-o_{l,m}^t) \big) \frac{\partial f_{\theta_{L+1}}}{ \partial o_{l, m}^t}
	= (2o_{l,m}^t - 1) \nabla_{o^t_{l,m}} f_{\theta_{L+1}}.
	\end{aligned}
\end{equation}
This approximation introduces bias to the gradient estimator, except when the map $f$ is multilinear \cite{tokui2017evaluating,shekhovtsov2020path}. By Equations \ref{general_estimator}, \ref{L_approx}, we have that 
\begin{equation} \label{eq:mid1}
	\hat{g}_l = \sum_m {\nabla_{\theta_l} p_{\theta_l} (o_{l,m}^t | \boldsymbol{o}_{l-1}^t, o_{l,m}^{t-1})} 
	{(2o_{l,m}^t - 1) 
	\nabla_{o^t_{l,m}} f_{\theta_{L+1}} }.
\end{equation}
By Equation \ref{eq:ot}, we have that ${\nabla_{\theta_l} p_{\theta_l} (o_{l,m}^t | \boldsymbol{o}_{l-1}^t, o_{l,m}^{t-1})} =(2o^t_{l,m}-1)F^\prime_\epsilon(u^t_{l,m}-v_{\text{th}})\nabla_{\theta_l}u^t_{l,m}$.
Therefore, by Equation \ref{eq:mid1}, we can formulate the noise-driven learning rule as \cite{fremaux2016neuromodulated,gerstner2018eligibility}
\begin{equation} \label{final1}
	\text{NDL:  }
	\hat{g}_l = 
	\sum_m 
	\underbrace{\nabla_{\theta_l} u^t_{l,m}}_{\text{ Pre-synaptic factor}}
	\overbrace{F_\epsilon^\prime (u_{l,m}^t -v_{\text{th}})}^{\text{ Post-synaptic factor}}
	\underbrace{\nabla_{o^t_{l,m}} f_{\theta_{L+1}}}_{\text{ Global learning signal}}.
\end{equation}
Computing synaptic weights update using Equation \ref{final1} does not require calculating an additional gradient generator in the forward pass, and $\hat{g}_l$ can be computed layer by layer in a single backward passage. Specifically, the gradient estimation is performed by a backward pass, where the $\nabla_{u_{l,m}^t} o^t_{l,m}$ term in the exact chain rule is replaced by a term computed using the noise probability density function $F^\prime_\epsilon$. Therefore, NDL is easy to implement and can mesh well with modern automatic differentiation frameworks. And, since NDL is backpropagation-compatible, we can use it to optimize NSNNs of any architecture easily. 

Interestingly, some previous works \cite{shrestha2018slayer} also constructed a surrogate gradient function (pseudo derivative to surrogate the $\nabla_{u_{l,m}^t} o^t_{l,m}$ term) by empirically adding infinitesimal Gaussian perturbations to the spiking neuron. The surrogate gradient function obtained by their single-neuron analysis shares similar insights as the post-synaptic term in NDL. However, our results are obtained from a network-level derivation. 

\subsection*{Theoretical analyses on internal noise and stability}

\begin{figure}
\centering
    \includegraphics[width=0.7 \textwidth]{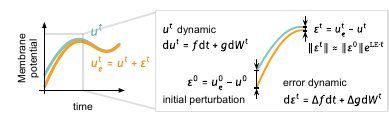}
    \caption{
    	{\bf Illustration of the theoretical analysis.} 
    	We study the stability of continuous NSNN sub-threshold dynamic (Equation \ref{eq:nlifnet}) given a small initial perturbation $\boldsymbol{\varepsilon}^0$ by analyzing the stability of the error dynamic (Equation \ref{eq:sde_error_sm}).  If the trivial solution $\boldsymbol{\varepsilon}^t=0$ (with $\mathtt{LE}$ being the sample Lyapunov exponent) of the error dynamic is stable. In this case, the NSNN system can self-correct small perturbations, keeping the membrane potential in a controllable range and thus producing stable outputs.
    }
    \label{fig:dynamicaltheo}
\end{figure}

Next, we analyze the stability of continuous NSNN sub-threshold dynamics (refer to Figure \ref{fig:dynamicaltheo}). Our analytical results show that, the internal noise in NSNNs can benefit the stability against small perturbations by allowing faster self-correction. This will also offer another lens through which to highlight the potential of NSNNs for improved performance and robustness \cite{basalyga2006response}. 
To preface the analyses, we model an NSNN layer as a special case of a stochastic system consisting of one drift term $\boldsymbol{f}$ (to model the deterministic part) and one diffusion term $\boldsymbol{g}$ (to model the stochastic part), formally,
\begin{equation} \label{eq:nlifnet}
	\mathrm{d}\boldsymbol{u}^t = \boldsymbol{f}(\boldsymbol{u}^t, \boldsymbol{I}^t)\mathrm{d}t + \boldsymbol{g}(\boldsymbol{u}^t, \boldsymbol{I}^t) \mathrm{d}\boldsymbol{W}^t, 
\end{equation}
where $\boldsymbol{u}^t,\boldsymbol{I}^t \in \mathbb{R}^{\dim(\boldsymbol{u}^t)}$, $\boldsymbol{W}^t$ is a $\dim(\boldsymbol{u}^t)$-dimensional Wiener process. And the coefficient $\boldsymbol{f}, \boldsymbol{g}$ satisfy the following assumption.
\begin{assumption} \label{as:lip}
	There exists a constant $K>0$ such that 
	$
	\Vert \boldsymbol{f}(\boldsymbol{u}^t_1,t) - \boldsymbol{f}(\boldsymbol{u}^t_2, t)\Vert + \Vert \boldsymbol{g}(\boldsymbol{u}^t_1,t)-\boldsymbol{g}(\boldsymbol{u}^t_2,t) \Vert \leq K \Vert \boldsymbol{u}^t_1 -\boldsymbol{u}^t_2 \Vert 
	$ for all $\boldsymbol{u}_1^t, \boldsymbol{u}^t_2 \in \mathbb{R}^{\dim(\boldsymbol{u}^t)}, t\in\mathbb{R}^+$.
\end{assumption}
In particular, we focus on the choice of $\boldsymbol{f}$ and $\boldsymbol{g}$:
\begin{equation} \label{eq:fng}
\begin{aligned}
	\boldsymbol{f}(\boldsymbol{u}^t, \boldsymbol{I}^t) &= a_1 \boldsymbol{u}^t + \boldsymbol{B}_1 \boldsymbol{I}^t,
	\\
	\boldsymbol{g}(\boldsymbol{u}^t, \boldsymbol{I}^t) &= {a_2 \operatorname{diag}(1)} + {b_2 \boldsymbol{f}(\boldsymbol{u}^t, \boldsymbol{I}^t)},
\end{aligned}	
\end{equation}
with constants $a_1\in \mathbb{R}$, $a_2, b_2 \in [0, \infty)$ and matrix $\boldsymbol{B}_1 \in \mathbb{R}^{\dim(\boldsymbol{u}^t) \times \dim (\boldsymbol{u}^t)}$. We can recover an NSNN layer consisting of Noisy LIF neurons described in the main text by letting $a_1=-{1}/{\tau_m}$, $a_2=\sigma$ and $b_2=0$.

Further, consider initializing the system represented by Equations \ref{eq:nlifnet},\ref{eq:fng} with two slightly different initializations $\boldsymbol{u}^0$, $\boldsymbol{u}^{0}_e\triangleq \boldsymbol{u}^0+\boldsymbol{\varepsilon}^0$, where $\boldsymbol{\varepsilon}^0\in \mathbb{R}^{\dim(\boldsymbol{u})}$ is the initial external perturbation (error) on membrane potentials. Then, the evolution of error $\boldsymbol{\varepsilon}^t = \boldsymbol{u}^t_e - \boldsymbol{u}^t$ satisfies the stochastic differential equation (SDE)
\begin{equation} \label{eq:sde_error_sm}
\begin{aligned}
	\mathrm{d}\boldsymbol{\varepsilon}^t
	= {\Delta\boldsymbol{f}(\boldsymbol{\varepsilon}^t) \mathrm{d}t} 
	+ {\Delta\boldsymbol{g}(\boldsymbol{\varepsilon}^t) \mathrm{d}t},
\end{aligned}
\end{equation}
where 
\begin{equation}
\begin{aligned}
	\Delta\boldsymbol{f}(\boldsymbol{\varepsilon}^t) &\triangleq \left(\boldsymbol{f}(\boldsymbol{u}^t + \boldsymbol{\varepsilon}^t, \boldsymbol{I}^t) - \boldsymbol{f}(\boldsymbol{u}^t,\boldsymbol{I}^t) \right), 
	\\
	\Delta\boldsymbol{g}(\boldsymbol{\varepsilon}^t) &\triangleq \left( \boldsymbol{g}(\boldsymbol{u}^t + \boldsymbol{\varepsilon}^t, \boldsymbol{I}^t) - \boldsymbol{g}(\boldsymbol{u}^t, \boldsymbol{I}^t) \right).
\end{aligned}
\end{equation}
Here we assume that the two random processes $\boldsymbol{u}^t, \boldsymbol{u}^t_e=\boldsymbol{u}^t+\boldsymbol{\varepsilon}^t$ are driven by the same Wiener process to make the subtraction operation valid. Since $\Delta\boldsymbol{f}(0)=0$ and $\Delta\boldsymbol{g}(0)=0$, $\boldsymbol{\varepsilon}^t=0$ admits a trivial solution for Equation \ref{eq:sde_error_sm}, whose uniqueness is guaranteed by Assumption \ref{as:lip} \cite{mao2007stochastic}. We focus on analyzing the stability of the trivial solution $\boldsymbol{\varepsilon}^t=0$; if it is stable, small initial perturbation $\boldsymbol{\varepsilon}^0\neq0$ will be reduced as the system evolves. This means that the system described by Equation \ref{eq:nlifnet} is self-correcting and can function reliably in the face of small perturbations or errors. Specifically, we focus on the {\it almost sure exponential stability} defined as follows.

\noindent {\bf Almost sure exponential stability \cite{mao2007stochastic}.}
The trivial solution $\boldsymbol{\varepsilon}^t = 0$ is {\it almost surely exponentially stable} if the sample Lyapunov exponent $\mathtt{LE}\triangleq\limsup_{t \rightarrow \infty} \frac{1}{t} \log \Vert \boldsymbol{\varepsilon}^t \Vert$ is almost surely negative for all $ \boldsymbol{\varepsilon}^0 \in \mathbb{R}^{\dim(\boldsymbol{u}^t)}$.

With $\mathtt{LE}$ being the sample Lyapunov exponent of the trivial solution, there exists a positive constant $C$ and a random variable $\tau\in[0,\infty)$ such that for all $t>\tau$, $\Vert\boldsymbol{\varepsilon}^t\Vert \leq C \exp(t\cdot\mathtt{LE})$ with probability $1$. Therefore, the almost sure exponential stability implies that almost all sample paths starting from non-zero initializations will tend to the equilibrium position $\boldsymbol{\varepsilon}^t = 0$ exponentially fast, i.e., the NSNN system can quickly self-correct small perturbations. Further, we have an important result regarding the bounds of the sample Lyapunov exponent.
\begin{theorem} \label{theo:bounds}
(Bounds for sample Lyapunov exponent of the trivial solution $\boldsymbol{\varepsilon}^t=0$). 
Suppose that $0\leq a_2\Vert \boldsymbol{\varepsilon}^t \Vert \leq \Vert \Delta \boldsymbol{g}(\boldsymbol{\varepsilon}^t) \Vert_F \leq (a_2+b_2) \Vert \boldsymbol{\varepsilon}^t \Vert$ for all non-zero $\boldsymbol{\varepsilon}^t \in \mathbb{R}^{\dim(\boldsymbol{u}^t)}$, $t \in \mathbb{R}^+$. Then, almost surely, 
$$
a_1-\frac{1}{2} a_2^2 - b_2^2 - 2 a_2 b_2 \leq \mathtt{LE} \leq a_1 - \frac{1}{2} a_2^2 + \frac{1}{2} b_2^2 + a_2 b_2
$$
for all $\boldsymbol{\varepsilon}^0 \in \mathbb{R}^{\dim (\boldsymbol{u}^t)}$. 
\end{theorem}

To prove Theorem 1, we first introduce the differential operator $L$. And for the brevity of notations, we temporally omit the superscript $t$, viz., we use $\boldsymbol{u}, \boldsymbol{\varepsilon}$ rather than $\boldsymbol{u}^t, \boldsymbol{\varepsilon}^t$ here. The differential operator associated with Equation \ref{eq:sde_error_sm} is defined by
\begin{equation}\label{eq:lll}
\begin{aligned}
    L = \frac{\partial }{\partial t}
    + \sum_{i=1}^{\dim(\boldsymbol{u})} \Delta \boldsymbol{f}_i (\boldsymbol{\varepsilon}, t) \frac{\partial }{\partial \boldsymbol{\varepsilon}_i}
    + \frac{1}{2} \sum_{i,j=1}^{\dim(\boldsymbol{u})} [\Delta \boldsymbol{g}^\top(\boldsymbol{\varepsilon}, t) \Delta \boldsymbol{g}(\boldsymbol{\varepsilon}, t)]_{ij} \frac{\partial^2}{\partial \boldsymbol{\varepsilon}_i \partial \boldsymbol{\varepsilon}_j}, 
\end{aligned}
\end{equation}
where the subscripts denote the entries of tensors, we also introduce a lemma as follows:
\begin{lemma} \label{lem:1}
(Stochastic Lyapunov theorem).
    Assume that there exists a non-negative real valued function $V(\boldsymbol{\varepsilon}, t)$ defined on $\mathbb{R}^{\dim(\boldsymbol{u})} \times \mathbb{R}^+$, denoted as $V \in C^{2,1} (\dim(\boldsymbol{u}) \times \mathbb{R}^+; \mathbb{R}^+)$. The function $V$ has continuous partial derivatives denoted as 
    \begin{equation} \nonumber
    \begin{aligned}
        V_{\boldsymbol{\varepsilon}} = \frac{\partial V}{\partial \boldsymbol{\varepsilon}},     
        V_t = \frac{\partial V}{\partial t},
        V_{\boldsymbol{\varepsilon}, \boldsymbol{\varepsilon}} = \frac{\partial^2 V}{\partial \boldsymbol{\varepsilon} \partial \boldsymbol{\varepsilon}^\top},
    \end{aligned}
    \end{equation}
    and constants $c_1, C_1>0$, $c_2, C_2 \in \mathbb{R}$, $c_3, C_3 \geq 0$ such that for all non-zero $\boldsymbol{\varepsilon}$ and $t\in \mathbb{R}^+$, 
    \begin{enumerate}
        \item $c_1 \Vert \boldsymbol{\varepsilon} \Vert^2 \leq V(\boldsymbol{\varepsilon}, t) \leq C_1\Vert \boldsymbol{\varepsilon} \Vert^2 $,
        \item $c_2 V(\boldsymbol{\varepsilon}, t) \leq LV(\boldsymbol{\varepsilon}, t) \leq C_2 V(\boldsymbol{\varepsilon}, t)$,
        \item $c_3 V(\boldsymbol{\varepsilon}, t)^2 \leq \Vert V_{\boldsymbol{\varepsilon}}(\boldsymbol{\varepsilon}, t) \Delta \boldsymbol{g}(\boldsymbol{\varepsilon}) \Vert^2_F \leq C_3 V(\boldsymbol{\varepsilon}, t)^2$. 
    \end{enumerate}
    Then, the sample Lyapunov exponent satisfies
    $$
    \frac{c_2}{2} - \frac{C_3}{4} \leq \limsup_{t \rightarrow \infty} \frac{1}{t} \log \Vert \boldsymbol{\varepsilon} \Vert \leq \frac{C_2}{2} - \frac{c_3}{4} 
    \ \ \ \ \text{a.s.}
    $$
\end{lemma}

This Lemma can be proved by combining Theorems 3.3 and 3.5 in ref. \cite{mao2007stochastic} Chapter 4 with $p=2$ as in ref. \cite{lim2021noisy}. With Lemma \ref{lem:1} in tow, we then prove Theorem 1.

\begin{proof} 
    Let $V(\boldsymbol{\varepsilon},t) = \Vert \boldsymbol{\varepsilon} \Vert^2$. If we assign $c_1=C_1=1$, condition 1 in Lemma \ref{lem:1} is satisfied. By Equation \ref{eq:lll}, we have 
    \begin{equation} \nonumber
    \begin{aligned}
    LV(\boldsymbol{\varepsilon},t) 
    =  V_{\boldsymbol{\varepsilon}}^\top \Delta\boldsymbol{f}(\boldsymbol{\varepsilon},t)
    + \frac{1}{2} \operatorname{trace}\left( \Delta\boldsymbol{g}^\top (\boldsymbol{\varepsilon}, t) V_{\boldsymbol{\varepsilon},\boldsymbol{\varepsilon}} \Delta\boldsymbol{g}(\boldsymbol{\varepsilon}, t)  \right) , 
    \end{aligned}
    \end{equation}
    Then, by Equations \ref{eq:fng},\ref{eq:sde_error_sm}, we can easily show that for all non-zero $\boldsymbol{\varepsilon} \in \mathbb{R}^{\dim(\boldsymbol{u})}$, $t \in \mathbb{R}^+$,
    $$
	    (2a_1 + a_2^2) \Vert \boldsymbol{\varepsilon} \Vert^2 
    	\leq LV(\boldsymbol{\varepsilon},t) 
    	\leq (2a_1+(a_2+b_2)^2)\Vert \boldsymbol{\varepsilon} \Vert^2 .
    $$
    Hence, condition 2 in Lemma \ref{lem:1} is satisfied. 
    
    To satisfy the condition 3 in Lemma \ref{lem:1}, by the assumption on $\Delta\boldsymbol{g}$, we have 
    $$
    4a_2^2 V^2 \leq \Vert V_{\boldsymbol{\varepsilon}}\Delta\boldsymbol{g} \Vert^2_F \leq 4(a_2+b_2)^2V^2.
    $$ 
    
    In conclusion, let $c_1=1,C_1=1$, $c_2=2a_1+a_2^2, C_2=2a_1+(a_2+b_2)^2$, $c_3=4a_2^2, C_3=4(a_2+b_2)^2$, we obtain the bounds of the sample Lyapunov exponent by Lemma \ref{lem:1}, that is 
    $$
	a_1-\frac{1}{2} a_2^2 - b_2^2 - 2 a_2 b_2 
	\leq \mathtt{LE} 
	\leq a_1 - \frac{1}{2} a_2^2 + \frac{1}{2} b_2^2 + a_2 b_2.
	$$
$\hfill\square$
\end{proof}

According to Theorem 1, the lower bound LB and upper bound UB of the sample Lyapunov exponent of the trivial solution $\boldsymbol{\varepsilon}^t = 0$ is $$LB=a_1-\frac{1}{2} a_2^2 - b_2^2 - 2 a_2 b_2, \\ UB=a_1 - \frac{1}{2} a_2^2 + \frac{1}{2} b_2^2 + a_2 b_2.$$ 
Generally, by the definition of almost sure exponential stability, if $UB<0$, the trivial solution is almost sure exponentially stable. And even if $a_1>0$, the system can still be stabilized by introducing internal noise terms to ensure a negative upper bound. For LIF-modeling spiking neural networks, as $a_1=-\frac{1}{\tau_m} < 0$, if $\frac{1}{\tau_m} > -\frac{1}{2} a_2^2 + \frac{1}{2} b_2^2 + a_2 b_2$, then we have that, the sub-threshold dynamic system is self-correcting, i.e., $\boldsymbol{\varepsilon}^t \overset{\text{a.s.}}{\rightarrow}0$. We then turn to the special case when the noise is additive $(a_2>0, b_2\rightarrow 0)$. NSNNs introduced in this article belong to this category. In the additive noise case, the multiplicative noise component in the system tends to zero $b_2 \rightarrow 0$, and the upper bound of the sample Lyapunov exponent becomes $UB_{\text{additive}}= a_1 - \frac{1}{2} a_2^2$. On the other hand, in the noiseless case, both noise terms tend to zero, $a_2\rightarrow0, b_2 \rightarrow 0$, and the lower bound is $LB_{\text{noise free}}=a_1$. Therefore we have that $UB_{\text{additive}} < LB_{\text{noise free}}=a_1 < 0$. In a small time interval, for the error at time $t$, we have $\Vert \boldsymbol{\varepsilon}^t \Vert = \Vert \boldsymbol{\varepsilon}^0 \Vert \exp(\mathtt{LE}\cdot t)$. Therefore, introducing additive noise ensures a more negative Lyapunov exponent than in the noiseless case, resulting in faster self-correction. This allows rapid detection and correction of factors that may cause instability, thereby improving NSNN system stability.

\subsection*{Experimental details}

 Experiments were conducted using a workstation with an Intel I5-10400, 64 GB RAM, and two NVIDIA RTX 3090s. The results are reported as mean and SD across multiple independent runs.
\subsubsection*{Details of recognition experiments}

The CIFAR dataset \cite{krizhevsky2009learning} includes 50k $32 \times 32$ images for training and 10k for evaluation. We adopt random crop, random horizontal flip, and AutoAugment \cite{cubuk2018autoaugment} for the training samples. The preprocessed samples are normalized using z-score scaling in the training and evaluation phases. The DVS-CIFAR dataset \cite{li2017cifar10} is a challenging neuromorphic benchmark recorded via a DVS camera using images from CIFAR-10. We adopt pre-processing pipeline following previous works \cite{samadzadeh2020convolutional}, i.e., divide the original set into a 9k-sample training set and 1k-sample evaluation set, and all event stream files are spatially downsampled to $48 \times 48$. We augment the training samples following previous studies \cite{deng2021temporal}. The DVS-Gesture dataset \cite{amir2017low} is recorded using a DVS128 event camera. It contains recordings of 11 hand gestures from 29 subjects under three illumination conditions.

We optimized SNN models using the dataset's observation-label pairs and calculated performance metrics on non-overlapping test data. Prediction accuracy was used as the performance metric for these tasks. For each network architecture and SNN algorithm combination, we set the optimal hyperparameters for DSNNs and NSNNs under different architecture and algorithm combinations via grid search to ensure a fair comparison. We indicate the corresponding simulation timestep in the results, which is the simulation duration of our discrete SNN implementation. The static image is repeatedly inputted for static image data, so a longer simulation timestep usually results in more accurate recognition. For event stream datasets, the fixed-length continuous event streams were discretized into simulation timestep time windows. The information embedded in those event streams will be spread over many time steps. Hence, a larger simulation timestep usually leads to more refined computation and more accurate recognition.

We set the standard deviation of membrane noise to 0.3 for CIFAR-10, CIFAR-100, and DVS-Gesture experiments and 0.2 for DVS-CIFAR ones. These configurations offer a fair balance between performance and resilience. For the SGL of DSNNs, we employ the ERF surrogate gradient $\text{SG}_{\text{ERF}}(x) = \frac{1}{\sqrt{\pi }} \exp(-x^2)$. Adam solvers \cite{kingma2014adam} with the cosine annealing learning rate scheduler \cite{loshchilov2016sgdr} was used to train all networks. We list hyper-parameters we adopted in recognition experiments in Table \ref{tab:7}. The initial learning rate is obtained through a grid search. 
The ResNet-19 \cite{zheng2021going}, VGGSNN \cite{deng2021temporal}, CIFARNet \cite{Wu2019}, and 7B-Net \cite{fang2021deep} SNN architectures used in this article follow the original implementation in previous works. The ResNet-18 architecture is given by 64C3-2(64C3-4C3)-2(128C3-128C3)-2(256C3-256C3)-2(512C3-512C3)-AP-FC. AP denotes average pooling, FC denotes the fully connected layer, and C denotes the convolution layer. The SNN algorithms STCA \cite{Gu2019}, STBP \cite{Wu2018}, STBP-tdBN \cite{zheng2021going} (tdBN), and TET \cite{deng2021temporal} used here follow the original implementations presented in these works. Some performances reported in the comparison come from the original paper, including those of LIAF-Net \cite{wu2021liaf}, STCA \cite{Gu2019}, STBP-tdBN \cite{zheng2021going} with ResNet-19 and ResNet-17, TET \cite{deng2021temporal} with ResNet-19, Wide-7B-Net \cite{fang2021deep}, and 7B-Net \cite{fang2021deep}. 

\begin{figure}
\centering
    \includegraphics[width=1 \textwidth]{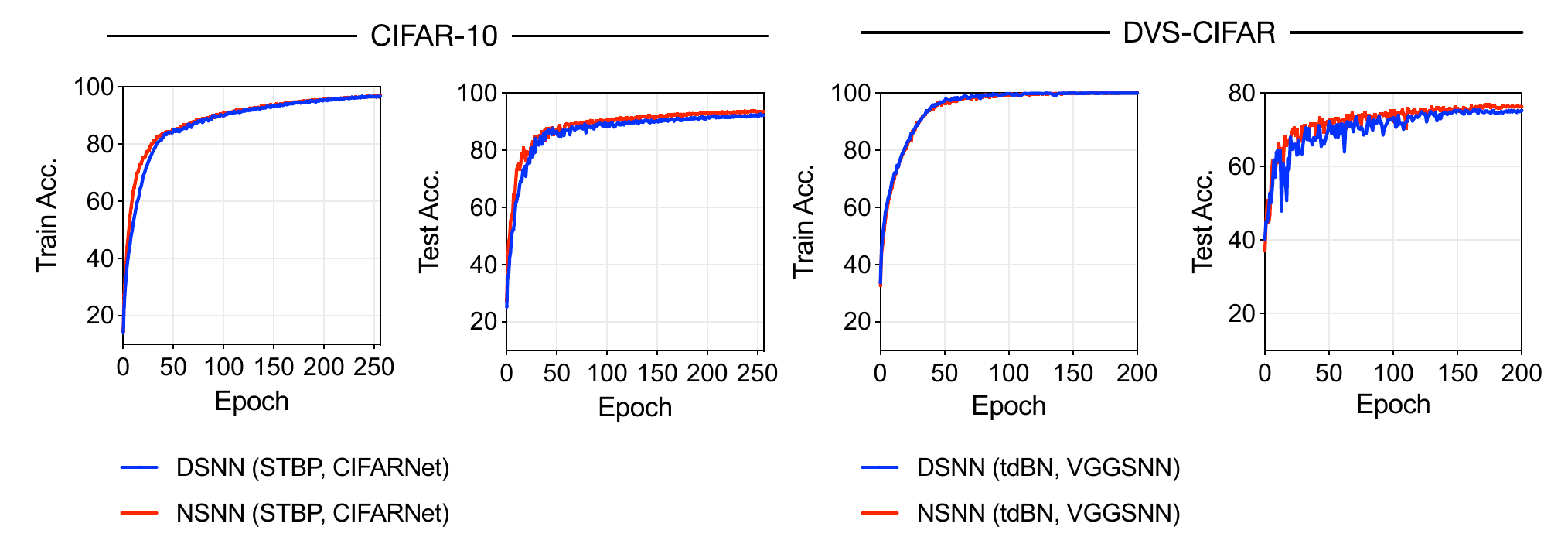}
    \caption{
    	{\bf Example learning curves on CIFAR-10 and DVS-CIFAR.}
   	}
    \label{fig:learn_curve}
\end{figure}

We visualize representative learning curves of DSNNs and NSNNs. As seen from Figure \ref{fig:learn_curve}, NSNNs demonstrated higher learning efficiency than DSNNs. And this advantage is more pronounced on datasets prone to overfitting, such as the DVS-CIFAR data. We also noticed that using NSNNs results in a slight increase in training time compared to DSNNs. This is mainly because the firing process in Noisy LIF neurons involves computing the firing probability and sampling from the Bernoulli distribution. Much of this additional time comes from the sampling step since the PyTorch framework we use has no targeted optimizations for sampling from random distributions. For instance, training a CIFARNet on CIFAR-10 data takes about 183 minutes for DSNNs and 194 minutes for NSNNs. 

\subsubsection*{Effect of internal noise level on network performance}
Although the noise-based computation and learning approaches were presented before, one question remains: how do we choose an appropriate noise scale for the Noisy Spiking Neural Networks?
The internal noise level influences learning as the post-synaptic factor $F^\prime_\epsilon(u-v_{\text{th}})$ in NDL is calculated using the probability density function of membrane noise $\epsilon$. When the variance is tiny, the noise distribution converges to a Dirac distribution with limited information (as measured by information entropy), preventing the post-synaptic factor from obtaining enough information to perform learning effectively. In the case of inference, the noise level directly influences the randomness of the firing distribution; smaller noise causes less corruption of the neuronal dynamics, while larger noise (with high variance) would over-corrupt the membrane voltage dynamics, therefore disrupting the flow of decisive information in the network, leading to greatly deteriorated performance. 

The previous content provided a theoretical explanation regarding internal noise, network stability, and performance. Here we show empirically that mild noise is beneficial for network performance. In early theoretical research \cite{basalyga2006response}, through the analysis of neuronal dynamics, researchers pointed out that under certain conditions, noise can have a positive impact on performance. More recent literature \cite{liu2021neuronal} studied the learning performance of small-scale spiking neural models with neuronal white noise under the STDP learning rule. It is found that appropriate levels of neuronal noise could benefit the model’s learning ability, which is consistent with the experimental results we obtained from our quantitative analysis in this section. 

\begin{figure}
\centering
    \includegraphics[width=0.75 \textwidth]{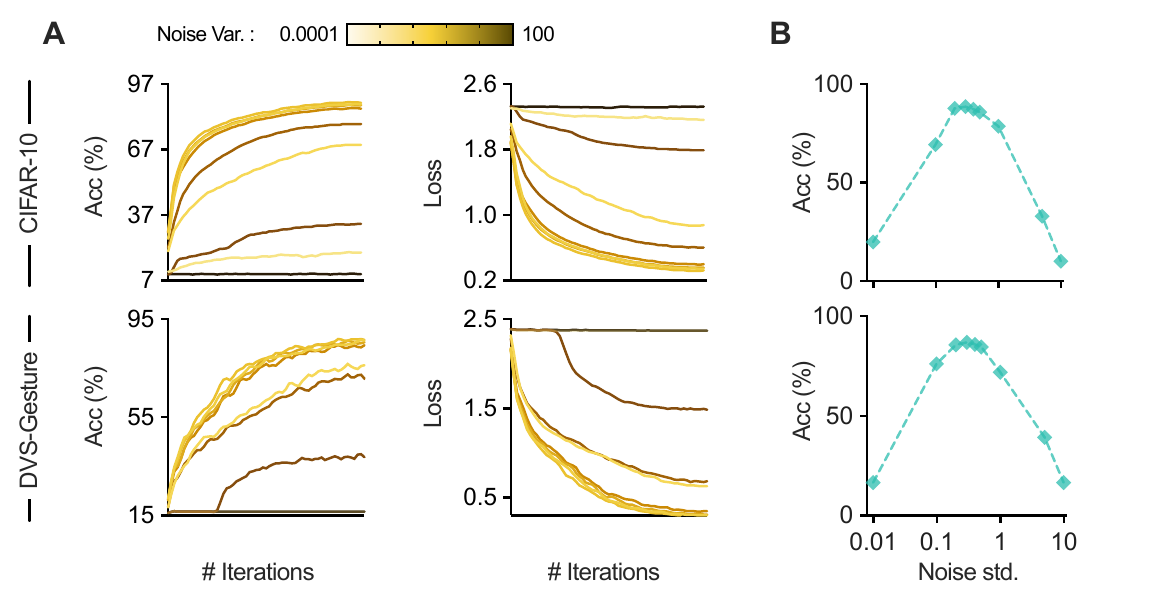}
    \caption{
    	{\bf Effect of internal noise level on performance.}
    	{\bf A} Learning curves of NSNNs under different noise levels, we use color to distinguish different noise levels.
    	{\bf B} The relationship between final test accuracy and the standard deviation of internal noise $\epsilon$. 
   	}
    \label{fig:noise_vs_perf}
\end{figure}

We ran experiments using the CIFAR-10 and DVS-Gesture datasets and trained identical networks with different internal noise level settings for 60 epochs. Results are presented by learning curves and the accuracy-standard deviation curves in Figure \ref{fig:noise_vs_perf}A. Our observations show that when the variance of internal noise increases from 0, the model’s performance initially improves and then declines. Notably, NSNNs achieve high performance near a moderate value (refer to Figure \ref{fig:noise_vs_perf}B), confirming our intuition that moderate noise is essential for high performance. According to our results, changes in $\text{std}[\epsilon]$ within a {\it moderate noise} range (from 0.2 to 0.5) have no significant effect on final performance. This gives us a range of internal noise levels to choose from when using NSNNs in practice. 

\subsubsection*{Details of recognition experiments with perturbations}

Here we list the details of the external perturbations in our experiments as follows. We denote the model to be evaluated as $\text{NN}$. In the Direct Optimization (DO) method, we construct the adversarial samples by directly solving the constrained optimization problem 
\begin{equation}
	\Delta x = \text{argmax}_{||\Delta x||_2 = \gamma } \text{loss} (f(x+\Delta x), y),	
\end{equation}
where $\Delta x$ for the adversarial perturbation and $x+\Delta x$ is the adversarial example. It is implemented using PyTorch and GeoTorch \cite{lezcano2019trivializations} toolkits. The L-2 norm bounded additive disturbance tensors are zero-initialized and optimized by an Adam solver with a learning rate of 0.002 for 30 iterations. After that, the additive perturbations are used to produce adversarial samples and fed into the target models (DSNNs or NSNNs in this work). 
The implementation of the FGSM method follows the original implementation \cite{goodfellow2014explaining}, where the adversarial example is constructed as 
\begin{equation}
	\tilde{x}^{adv} = x + \gamma_{FGSM} \times \text{sign}[ \nabla_{x} \text{loss} \left(\text{NN}(x), y) \right) ].	
\end{equation}

The input-level EventDrop \cite{gu2021eventdrop} perturbation for dynamic inputs is constructed by randomly dropping spikes in the raw input spike trains. The dropping probability is set by a parameter $\rho$. The strategy of dropping we consider is Random Drop \cite{gu2021eventdrop}, which combines spatial and temporal-wise event-dropping strategies. During the evaluation, we first individually performed EventDrop over every sample from the test set and then fed our testees with the disturbed inputs.

The spike state-level perturbation includes two disturbances: the emission state from 1 to 0 (spike to silence) and the emission state from 0 to 1 (silence to spike). To simplify the settings, we use one parameter $\beta$, to control the probability of both kinds of disturbances. Let variable $o_{\text{old}}$ denotes the original spike state, if $o_{\text{old}}=1$, we have $\mathbb{P}[o_{\text{new}} = 0] = \beta$, else, if $o_{\text{old}}=0$, $\mathbb{P}[o_{\text{new}} = 1] = \beta$. 

\subsubsection*{Details of recognition task coding analyses} 

In this part, we used the learned models in previous recognition experiments, whose simulation timesteps are 2, 2, 10, and 16, respectively. The number of test samples for computing the Pearson correlation coefficients is 500 for CIFAR-10, CIFAR-100, DVS-CIFAR, and 200 for DVS-Gesture. 
In our analyses, the spike count variability is measured by the Fano factor \citep{fano1947ionization,edmund2010}, which is calculated as follows. Let $n^{\text{trial ID}}_{L,m}$ be the spike count of neuron $(L,m)$, and the mean value is $\overline{n}_{L,m} = \frac{1}{\text{\# trials}} \sum_{k} n_{L,m}^k$, the deviations from that mean is computed as $\Delta n_{L,m}^{\text{trial ID}} = n_{L,m}^{\text{trial ID}} - \overline{n}_{L,m}$, then, Fano Factor is given by $\text{FF}_{L,m} = \frac{\text{Var}[n_{L,m}]}{\overline{n}_{L,m}}$.

\subsubsection*{Details of neural activity fitting experiments}

In the neural activity fitting experiments, we used neural recordings of dark-adapted axolotl salamander retinas \cite{onken2016using}. The original dataset contains the retinal neural responses of two retinas to two movies. We only used a part of the data (responses of retina 2 to movie 2) in this experiment. It contains complex natural scenes of a tiger on a prey hunt, roughly 60 $s$ long. The movie was discretized into bins of 33 $ms$, and all frames were converted to grayscale with a resolution of 360 pixel$\times$360 pixel at 7.5 $\mu m$$\times$7.5 $\mu m$ per pixel, covering a 2700 $\mu m\times$2700 $\mu m$ area on the retina. The neural recording contains 42 repetitions of 49 cells. We partitioned all records into stimuli-response sample pairs of 1 second and down-sampled the frames to 90 pixel$\times$90 pixel. The data was split into non-overlapping train/test (50\%/50\%) parts. 


The network architecture for the DSNN and NSNN models is 16C25-32C11-FC64-FC64-FC64-FC49. FC denotes the fully connected layer, and C denotes the convolutional layer. These models were trained using the Adam optimizer with a cosine-decay learning rate scheduler, starting at a rate of 0.0003. The mini-batch size was set to 64, and the models were trained for 64 epochs. At test time, the test samples have the same length as the training samples (1 second) by default. During training, the DSNN and NSNN models are optimized to minimize the Maximum Mean Discrepancy (MMD) loss \cite{park2013kernel,arribas2020rescuing}. We use the first-order postsynaptic potential (PSP) kernel \cite{zenke2018superspike} that can effectively depict the temporal dependencies in spike train data. Denoting the predicted, recorded spike trains as $\hat{\mathbf{y}}$, $\mathbf{y}$, respectively, we can write the PSP kernel MMD predictive loss as 
\begin{equation}
	\mathcal{L}_{\text{PSP-MMD}}
	= 
	\frac{1}{T} \sum_{t=1}^T
	\sum_{\tau=1}^t
	 \big\Vert 
	 	\mathrm{PSP}(\hat{\mathbf{y}}_{1:\tau})
	 	- 
	 	\mathrm{PSP}(\mathbf{y}_{1:\tau})
	 \big\Vert^2, 
\end{equation}
where $\mathrm{PSP}(\mathbf{y}_{1:\tau}) = (1-\frac{1}{\tau_s}) \mathrm{PSP} (\mathbf{y}_{1:\tau-1}) + \frac{1}{\tau_s} \mathbf{y}_\tau$, and we set the time constant $\tau_s=2$. For the DSNN model, we used the ERF surrogate gradient as in the recognition experiments. And for the NSNN model, the internal noise is $\mathcal{N}(0, 0.2^2)$.  

The CNN baseline we used adopts the architecture: 32C25-BN-16C11-BN-FC49-BN-PSoftPlus, where BN denotes the Batch-Normalization operation, and PSoftPlus denotes the parameterized SoftPlus activation. The training specifications follow the implementation in the previous work \cite{mcintosh2016}. 




\section*{Acknowledgments}

This work was supported by National Key Research and Development Program of China under Grant No. 2020AAA0105900 and National Natural Science Foundation of China under Grant No. 62236007 and No. 62276235. 

We thank Dr. Seiya Tokui and Prof. Penghang Yin for their insightful discussions, as well as to the reviewers for their constructive suggestions.

\section*{Author contributions}

Conceptualization, G.M., R.Y., and H.T.; Formal analysis, G.M.; Investigation, G.M.; Methodology and software, G.M.; Writing - original draft, G.M.; Visualization, G.M.; Writing - review and editing, G.M., R.Y., and H.T.; Funding acquisition, H.T., and R.Y.; Supervision, H.T., and R.Y.

\section*{Declaration of interests}

The authors declare no competing interests.

\section*{Inclusion and diversity statement}

While citing references scientifically relevant for this work, we also actively worked to promote gender balance in our reference list.

\clearpage 
\bibliography{new_references.bib}

\begin{thebibliography}{119}
\providecommand{\natexlab}[1]{#1}
\providecommand{\url}[1]{\texttt{#1}}
\providecommand{\href}[2]{#2}
\providecommand{\path}[1]{#1}
\providecommand{\DOIprefix}{doi:}
\providecommand{\ArXivprefix}{arXiv:}
\providecommand{\URLprefix}{URL: }
\providecommand{\Pubmedprefix}{pmid:}
\providecommand{\doi}[1]{\href{http://dx.doi.org/#1}{\path{#1}}}
\providecommand{\Pubmed}[1]{\href{pmid:#1}{\path{#1}}}
\providecommand{\BIBand}{and}
\providecommand{\bibinfo}[2]{#2}
\ifx\xfnm\undefined \def\xfnm[#1]{\unskip,\space#1}\fi
\makeatletter\def\@biblabel#1{#1.}\makeatother
\bibitem[{Zilli and Hasselmo(2010)}]{zilli2010coupled}
\bibinfo{author}{Zilli, E.~A.}, and \bibinfo{author}{Hasselmo, M.~E.}
  (\bibinfo{year}{2010}). \bibinfo{title}{Coupled noisy spiking neurons as
  velocity-controlled oscillators in a model of grid cell spatial firing}.
\newblock \bibinfo{journal}{Journal of Neuroscience}
  \emph{\bibinfo{volume}{30}}, \bibinfo{pages}{13850--13860}.
  \DOIprefix\doi{https://doi.org/10.1523/JNEUROSCI.0547-10.2010}.
\bibitem[{Teeter et~al.(2018)Teeter, Iyer, Menon, Gouwens, Feng, Berg, Szafer,
  Cain, Zeng, Hawrylycz et~al.}]{teeter2018generalized}
\bibinfo{author}{Teeter, C.}, \bibinfo{author}{Iyer, R.},
  \bibinfo{author}{Menon, V.}, \bibinfo{author}{Gouwens, N.},
  \bibinfo{author}{Feng, D.}, \bibinfo{author}{Berg, J.},
  \bibinfo{author}{Szafer, A.}, \bibinfo{author}{Cain, N.},
  \bibinfo{author}{Zeng, H.}, \bibinfo{author}{Hawrylycz, M.} et~al.
  (\bibinfo{year}{2018}). \bibinfo{title}{Generalized leaky integrate-and-fire
  models classify multiple neuron types}.
\newblock \bibinfo{journal}{Nature Communications} \emph{\bibinfo{volume}{9}},
  \bibinfo{pages}{709}.
  \DOIprefix\doi{https://doi.org/10.1038/s41467-017-02717-4}.
\bibitem[{Simonyan and Zisserman(2014)}]{simonyan2014very}
\bibinfo{author}{Simonyan, K.}, and \bibinfo{author}{Zisserman, A.}
  (\bibinfo{year}{2014}). \bibinfo{title}{Very deep convolutional networks for
  large-scale image recognition}.
\newblock \bibinfo{journal}{arXiv}.
  \DOIprefix\doi{https://doi.org/10.48550/arXiv.1409.1556}.
\bibitem[{Szegedy et~al.(2015)Szegedy, Liu, Jia, Sermanet, Reed, Anguelov,
  Erhan, Vanhoucke and Rabinovich}]{szegedy2015going}
\bibinfo{author}{Szegedy, C.}, \bibinfo{author}{Liu, W.}, \bibinfo{author}{Jia,
  Y.}, \bibinfo{author}{Sermanet, P.}, \bibinfo{author}{Reed, S.},
  \bibinfo{author}{Anguelov, D.}, \bibinfo{author}{Erhan, D.},
  \bibinfo{author}{Vanhoucke, V.}, and \bibinfo{author}{Rabinovich, A.}
\newblock \bibinfo{title}{Going deeper with convolutions}.
\newblock In: \emph{\bibinfo{booktitle}{The IEEE / CVF Computer Vision and
  Pattern Recognition Conference}} (\bibinfo{year}{2015}):\unskip(
  \bibinfo{pages}{1--9}).
\newblock \DOIprefix\doi{https://doi.org/10.1109/CVPR.2015.7298594}.
\bibitem[{He et~al.(2016)He, Zhang, Ren and Sun}]{he2016deep}
\bibinfo{author}{He, K.}, \bibinfo{author}{Zhang, X.}, \bibinfo{author}{Ren,
  S.}, and \bibinfo{author}{Sun, J.}
\newblock \bibinfo{title}{Deep residual learning for image recognition}.
\newblock In: \emph{\bibinfo{booktitle}{The IEEE / CVF Computer Vision and
  Pattern Recognition Conference}} (\bibinfo{year}{2016}):\unskip(
  \bibinfo{pages}{770--778}).
\newblock \DOIprefix\doi{https://doi.org/10.1109/CVPR.2016.90}.
\bibitem[{Lee et~al.(2016)Lee, Delbruck and Pfeiffer}]{Lee2016}
\bibinfo{author}{Lee, J.~H.}, \bibinfo{author}{Delbruck, T.}, and
  \bibinfo{author}{Pfeiffer, M.} (\bibinfo{year}{2016}).
  \bibinfo{title}{Training deep spiking neural networks using backpropagation}.
\newblock \bibinfo{journal}{Frontiers in Neuroscience}
  \emph{\bibinfo{volume}{10}}.
  \DOIprefix\doi{https://doi.org/10.3389/fnins.2016.00508}.
\bibitem[{Wu et~al.(2018)Wu, Deng, Li, Zhu and Shi}]{Wu2018}
\bibinfo{author}{Wu, Y.}, \bibinfo{author}{Deng, L.}, \bibinfo{author}{Li, G.},
  \bibinfo{author}{Zhu, J.}, and \bibinfo{author}{Shi, L.}
  (\bibinfo{year}{2018}). \bibinfo{title}{{Spatio-temporal backpropagation for
  training high-performance spiking neural networks}}.
\newblock \bibinfo{journal}{Frontiers in Neuroscience}
  \emph{\bibinfo{volume}{12}}, \bibinfo{pages}{1--12}.
  \DOIprefix\doi{https://doi.org/10.3389/fnins.2018.00331}.
\bibitem[{Wu et~al.(2019)Wu, Deng, Li, Zhu, Xie and Shi}]{Wu2019}
\bibinfo{author}{Wu, Y.}, \bibinfo{author}{Deng, L.}, \bibinfo{author}{Li, G.},
  \bibinfo{author}{Zhu, J.}, \bibinfo{author}{Xie, Y.}, and
  \bibinfo{author}{Shi, L.}
\newblock \bibinfo{title}{Direct training for spiking neural networks: Faster,
  larger, better}.
\newblock In: \emph{\bibinfo{booktitle}{Proceedings of the AAAI conference on
  artificial intelligence}}
  (\bibinfo{year}{2019}):\unskip\DOIprefix\doi{https://doi.org/10.1609/aaai.v33i01.33011311}.
\bibitem[{Deng et~al.(2021)Deng, Li, Zhang and Gu}]{deng2021temporal}
\bibinfo{author}{Deng, S.}, \bibinfo{author}{Li, Y.}, \bibinfo{author}{Zhang,
  S.}, and \bibinfo{author}{Gu, S.}
\newblock \bibinfo{title}{Temporal efficient training of spiking neural network
  via gradient re-weighting}.
\newblock In: \emph{\bibinfo{booktitle}{International Conference on Learning
  Representations}}
  (\bibinfo{year}{2021}):\unskip\DOIprefix\doi{https://doi.org/10.48550/arXiv.2202.11946}.
\bibitem[{Volinski et~al.(2022)Volinski, Zaidel, Shalumov, DeWolf, Supic and
  Tsur}]{volinski2022data}
\bibinfo{author}{Volinski, A.}, \bibinfo{author}{Zaidel, Y.},
  \bibinfo{author}{Shalumov, A.}, \bibinfo{author}{DeWolf, T.},
  \bibinfo{author}{Supic, L.}, and \bibinfo{author}{Tsur, E.~E.}
  (\bibinfo{year}{2022}). \bibinfo{title}{Data-driven artificial and spiking
  neural networks for inverse kinematics in neurorobotics}.
\newblock \bibinfo{journal}{Patterns} \emph{\bibinfo{volume}{3}},
  \bibinfo{pages}{100391}.
  \DOIprefix\doi{https://doi.org/10.1016/j.patter.2021.100391}.
\bibitem[{Zhao et~al.(2022)Zhao, Zeng, Han, Fang and Zhao}]{zhao2022nature}
\bibinfo{author}{Zhao, F.}, \bibinfo{author}{Zeng, Y.}, \bibinfo{author}{Han,
  B.}, \bibinfo{author}{Fang, H.}, and \bibinfo{author}{Zhao, Z.}
  (\bibinfo{year}{2022}). \bibinfo{title}{Nature-inspired self-organizing
  collision avoidance for drone swarm based on reward-modulated spiking neural
  network}.
\newblock \bibinfo{journal}{Patterns} \emph{\bibinfo{volume}{3}},
  \bibinfo{pages}{100611}.
  \DOIprefix\doi{https://doi.org/10.1016/j.patter.2022.100611}.
\bibitem[{Fang et~al.(2021)Fang, Yu, Chen, Huang, Masquelier and
  Tian}]{fang2021deep}
\bibinfo{author}{Fang, W.}, \bibinfo{author}{Yu, Z.}, \bibinfo{author}{Chen,
  Y.}, \bibinfo{author}{Huang, T.}, \bibinfo{author}{Masquelier, T.}, and
  \bibinfo{author}{Tian, Y.}
\newblock \bibinfo{title}{Deep residual learning in spiking neural networks}.
\newblock In: \emph{\bibinfo{booktitle}{Advances in Neural Information
  Processing Systems (NeurIPS)}}
  (\bibinfo{year}{2021}):\unskip\DOIprefix\doi{https://doi.org/10.48550/arXiv.2102.04159}.
\bibitem[{Liu et~al.(2020{\natexlab{a}})Liu, Ruan, Xing, Tang and
  Pan}]{Liu2020}
\bibinfo{author}{Liu, Q.}, \bibinfo{author}{Ruan, H.}, \bibinfo{author}{Xing,
  D.}, \bibinfo{author}{Tang, H.}, and \bibinfo{author}{Pan, G.}
\newblock \bibinfo{title}{Effective aer object classification using segmented
  probability-maximization learning in spiking neural networks}.
\newblock In: \emph{\bibinfo{booktitle}{Proceedings of the AAAI conference on
  artificial intelligence}}
  (\bibinfo{year}{2020}{\natexlab{a}}):\unskip\DOIprefix\doi{https://doi.org/10.1609/aaai.v34i02.5486}.
\bibitem[{Roy et~al.(2019)Roy, Jaiswal and Panda}]{roy2019towards}
\bibinfo{author}{Roy, K.}, \bibinfo{author}{Jaiswal, A.}, and
  \bibinfo{author}{Panda, P.} (\bibinfo{year}{2019}). \bibinfo{title}{Towards
  spike-based machine intelligence with neuromorphic computing}.
\newblock \bibinfo{journal}{Nature} \emph{\bibinfo{volume}{575}},
  \bibinfo{pages}{607--617}.
  \DOIprefix\doi{https://doi.org/10.1038/s41586-019-1677-2}.
\bibitem[{Verveen and DeFelice(1974)}]{verveen1974membrane}
\bibinfo{author}{Verveen, A.}, and \bibinfo{author}{DeFelice, L.}
  (\bibinfo{year}{1974}). \bibinfo{title}{Membrane noise}.
\newblock \bibinfo{journal}{Progress in Biophysics and Molecular Biology}
  \emph{\bibinfo{volume}{28}}, \bibinfo{pages}{189--265}.
  \DOIprefix\doi{https://doi.org/10.1016/0079-6107(74)90019-4}.
\bibitem[{Kempter et~al.(1998)Kempter, Gerstner, Van~Hemmen and
  Wagner}]{kempter1998extracting}
\bibinfo{author}{Kempter, R.}, \bibinfo{author}{Gerstner, W.},
  \bibinfo{author}{Van~Hemmen, J.~L.}, and \bibinfo{author}{Wagner, H.}
  (\bibinfo{year}{1998}). \bibinfo{title}{Extracting oscillations: Neuronal
  coincidence detection with noisy periodic spike input}.
\newblock \bibinfo{journal}{Neural computation} \emph{\bibinfo{volume}{10}},
  \bibinfo{pages}{1987--2017}.
  \DOIprefix\doi{https://doi.org/10.1162/089976698300016945}.
\bibitem[{Stein(1965)}]{stein1965theoretical}
\bibinfo{author}{Stein, R.~B.} (\bibinfo{year}{1965}). \bibinfo{title}{A
  theoretical analysis of neuronal variability}.
\newblock \bibinfo{journal}{Biophysical Journal} \emph{\bibinfo{volume}{5}},
  \bibinfo{pages}{173--194}.
  \DOIprefix\doi{https://doi.org/10.1016/s0006-3495(65)86709-1}.
\bibitem[{Stein et~al.(2005)Stein, Gossen and Jones}]{stein2005neuronal}
\bibinfo{author}{Stein, R.~B.}, \bibinfo{author}{Gossen, E.~R.}, and
  \bibinfo{author}{Jones, K.~E.} (\bibinfo{year}{2005}).
  \bibinfo{title}{Neuronal variability: noise or part of the signal?}
\newblock \bibinfo{journal}{Nature Reviews Neuroscience}
  \emph{\bibinfo{volume}{6}}, \bibinfo{pages}{389--397}.
  \DOIprefix\doi{https://doi.org/10.1038/nrn1668}.
\bibitem[{Faisal et~al.(2008)Faisal, Selen and Wolpert}]{faisal2008noise}
\bibinfo{author}{Faisal, A.~A.}, \bibinfo{author}{Selen, L.~P.}, and
  \bibinfo{author}{Wolpert, D.~M.} (\bibinfo{year}{2008}).
  \bibinfo{title}{Noise in the nervous system}.
\newblock \bibinfo{journal}{Nature Reviews Neuroscience}
  \emph{\bibinfo{volume}{9}}, \bibinfo{pages}{292--303}.
  \DOIprefix\doi{https://doi.org/10.1038/nrn2258}.
\bibitem[{Maass(1995)}]{maass1995noisy}
\bibinfo{author}{Maass, W.}
\newblock \bibinfo{title}{On the computational power of noisy spiking neurons}.
\newblock In: \emph{\bibinfo{booktitle}{Advances in Neural Information
  Processing Systems (NeurIPS)}} (\bibinfo{year}{1995}):\unskip(
  \bibinfo{pages}{211–217}).
\newblock \DOIprefix\doi{https://dl.acm.org/doi/abs/10.5555/2998828.2998858}.
\bibitem[{Maass(1996)}]{maass1996noisy}
\bibinfo{author}{Maass, W.}
\newblock \bibinfo{title}{Noisy spiking neurons with temporal coding have more
  computational power than sigmoidal neurons}.
\newblock In: \emph{\bibinfo{booktitle}{Advances in Neural Information
  Processing Systems (NeurIPS)}} (\bibinfo{year}{1996}):\unskip(
  \bibinfo{pages}{211–217}).
\newblock \DOIprefix\doi{https://dl.acm.org/doi/10.5555/2998981.2999011}.
\bibitem[{Patel and Kosko(2005)}]{patel2005stochastic}
\bibinfo{author}{Patel, A.}, and \bibinfo{author}{Kosko, B.}
  (\bibinfo{year}{2005}). \bibinfo{title}{Stochastic resonance in noisy spiking
  retinal and sensory neuron models}.
\newblock \bibinfo{journal}{Neural Networks} \emph{\bibinfo{volume}{18}},
  \bibinfo{pages}{467--478}.
  \DOIprefix\doi{https://doi.org/10.1016/j.neunet.2005.06.031}.
\bibitem[{Liu et~al.(2020{\natexlab{b}})Liu, Xiao, Si, Cao, Kumar and
  Hsieh}]{liu2020does}
\bibinfo{author}{Liu, X.}, \bibinfo{author}{Xiao, T.}, \bibinfo{author}{Si,
  S.}, \bibinfo{author}{Cao, Q.}, \bibinfo{author}{Kumar, S.}, and
  \bibinfo{author}{Hsieh, C.-J.}
\newblock \bibinfo{title}{How does noise help robustness? explanation and
  exploration under the neural sde framework}.
\newblock In: \emph{\bibinfo{booktitle}{The IEEE / CVF Computer Vision and
  Pattern Recognition Conference}}
  (\bibinfo{year}{2020}{\natexlab{b}}):\unskip( \bibinfo{pages}{282--290}).
\newblock \DOIprefix\doi{https://doi.org/10.1109/CVPR42600.2020.00036}.
\bibitem[{Camuto et~al.(2020)Camuto, Willetts, Simsekli, Roberts and
  Holmes}]{camuto2020explicit}
\bibinfo{author}{Camuto, A.}, \bibinfo{author}{Willetts, M.},
  \bibinfo{author}{Simsekli, U.}, \bibinfo{author}{Roberts, S.~J.}, and
  \bibinfo{author}{Holmes, C.~C.}
\newblock \bibinfo{title}{Explicit regularisation in gaussian noise
  injections}.
\newblock In: \emph{\bibinfo{booktitle}{Advances in Neural Information
  Processing Systems (NeurIPS)}} vol.~\bibinfo{volume}{33}
  (\bibinfo{year}{2020}):\unskip( \bibinfo{pages}{16603--16614}).
\newblock \DOIprefix\doi{https://dl.acm.org/doi/abs/10.5555/3495724.3497117}.
\bibitem[{Lim et~al.(2021)Lim, Erichson, Hodgkinson and Mahoney}]{lim2021noisy}
\bibinfo{author}{Lim, S.~H.}, \bibinfo{author}{Erichson, N.~B.},
  \bibinfo{author}{Hodgkinson, L.}, and \bibinfo{author}{Mahoney, M.~W.}
\newblock \bibinfo{title}{Noisy recurrent neural networks}.
\newblock In: \emph{\bibinfo{booktitle}{Advances in Neural Information
  Processing Systems (NeurIPS)}}
  (\bibinfo{year}{2021}):\unskip\DOIprefix\doi{https://doi.org/10.48550/arXiv.2102.04877}.
\bibitem[{Hinton et~al.(2012)Hinton, Srivastava, Krizhevsky, Sutskever and
  Salakhutdinov}]{hinton2012improving}
\bibinfo{author}{Hinton, G.~E.}, \bibinfo{author}{Srivastava, N.},
  \bibinfo{author}{Krizhevsky, A.}, \bibinfo{author}{Sutskever, I.}, and
  \bibinfo{author}{Salakhutdinov, R.~R.} (\bibinfo{year}{2012}).
  \bibinfo{title}{Improving neural networks by preventing co-adaptation of
  feature detectors}.
\newblock \bibinfo{journal}{arXiv}.
  \DOIprefix\doi{https://doi.org/10.48550/arXiv.1207.0580}.
\bibitem[{Gerstein and Mandelbrot(1964)}]{gerstein1964random}
\bibinfo{author}{Gerstein, G.~L.}, and \bibinfo{author}{Mandelbrot, B.}
  (\bibinfo{year}{1964}). \bibinfo{title}{Random walk models for the spike
  activity of a single neuron}.
\newblock \bibinfo{journal}{Biophysical Journal} \emph{\bibinfo{volume}{4}},
  \bibinfo{pages}{41--68}.
  \DOIprefix\doi{https://doi.org/10.1016/s0006-3495(64)86768-0}.
\bibitem[{Tuckwell(1989)}]{tuckwell1989stochastic}
\bibinfo{author}{Tuckwell, H.~C.}
\newblock \bibinfo{title}{Stochastic Processes in the Neurosciences}.
\newblock \bibinfo{publisher}{SIAM} (\bibinfo{year}{1989}).
\newblock \DOIprefix\doi{https://doi.org/10.1137/1032064}.
\bibitem[{Plesser and Gerstner(2000{\natexlab{a}})}]{plesser2000noise}
\bibinfo{author}{Plesser, H.~E.}, and \bibinfo{author}{Gerstner, W.}
  (\bibinfo{year}{2000}{\natexlab{a}}). \bibinfo{title}{Noise in
  integrate-and-fire neurons: from stochastic input to escape rates}.
\newblock \bibinfo{journal}{Neural Computation} \emph{\bibinfo{volume}{12}},
  \bibinfo{pages}{367--384}.
  \DOIprefix\doi{https://doi.org/10.1162/089976600300015835}.
\bibitem[{Gerstner et~al.(2014)Gerstner, Kistler, Naud and
  Paninski}]{gerstner2014neuronal}
\bibinfo{author}{Gerstner, W.}, \bibinfo{author}{Kistler, W.~M.},
  \bibinfo{author}{Naud, R.}, and \bibinfo{author}{Paninski, L.}
\newblock \bibinfo{title}{Neuronal dynamics: From single neurons to networks
  and models of cognition}.
\newblock \bibinfo{publisher}{Cambridge University Press}
  (\bibinfo{year}{2014}).
\newblock \DOIprefix\doi{https://doi.org/10.1017/CBO9781107447615}.
\bibitem[{Rao(2004{\natexlab{a}})}]{rao2004bayesian}
\bibinfo{author}{Rao, R.~P.} (\bibinfo{year}{2004}{\natexlab{a}}).
  \bibinfo{title}{Bayesian computation in recurrent neural circuits}.
\newblock \bibinfo{journal}{Neural Computation} \emph{\bibinfo{volume}{16}},
  \bibinfo{pages}{1--38}.
  \DOIprefix\doi{https://doi.org/10.1162/08997660460733976}.
\bibitem[{Rao(2004{\natexlab{b}})}]{rao2004hierarchical}
\bibinfo{author}{Rao, R.~P.}
\newblock \bibinfo{title}{Hierarchical bayesian inference in networks of
  spiking neurons}.
\newblock In: \emph{\bibinfo{booktitle}{Advances in Neural Information
  Processing Systems (NeurIPS)}} vol.~\bibinfo{volume}{17}
  (\bibinfo{year}{2004}{\natexlab{b}}):\unskip\DOIprefix\doi{https://dl.acm.org/doi/10.5555/2976040.2976180}.
\bibitem[{Deneve(2004)}]{deneve2004bayesian}
\bibinfo{author}{Deneve, S.}
\newblock \bibinfo{title}{Bayesian inference in spiking neurons}.
\newblock In: \emph{\bibinfo{booktitle}{Advances in Neural Information
  Processing Systems (NeurIPS)}} vol.~\bibinfo{volume}{17}
  (\bibinfo{year}{2004}):\unskip\DOIprefix\doi{https://dl.acm.org/doi/abs/10.5555/2976040.2976085}.
\bibitem[{Kasabov(2010)}]{kasabov2010spike}
\bibinfo{author}{Kasabov, N.} (\bibinfo{year}{2010}). \bibinfo{title}{To spike
  or not to spike: A probabilistic spiking neuron model}.
\newblock \bibinfo{journal}{Neural Networks} \emph{\bibinfo{volume}{23}},
  \bibinfo{pages}{16--19}.
  \DOIprefix\doi{https://doi.org/10.1016/j.neunet.2009.08.010}.
\bibitem[{Skatchkovsky et~al.(2021)Skatchkovsky, Jang and
  Simeone}]{skatchkovsky2021spiking}
\bibinfo{author}{Skatchkovsky, N.}, \bibinfo{author}{Jang, H.}, and
  \bibinfo{author}{Simeone, O.} (\bibinfo{year}{2021}). \bibinfo{title}{Spiking
  neural networks—part ii: Detecting spatio-temporal patterns}.
\newblock \bibinfo{journal}{IEEE Communications Letters}
  \emph{\bibinfo{volume}{25}}, \bibinfo{pages}{1741--1745}.
  \DOIprefix\doi{https://doi.org/10.1109/LCOMM.2021.3050242}.
\bibitem[{Neftci et~al.(2019)Neftci, Mostafa and Zenke}]{neftci2019surrogate}
\bibinfo{author}{Neftci, E.~O.}, \bibinfo{author}{Mostafa, H.}, and
  \bibinfo{author}{Zenke, F.} (\bibinfo{year}{2019}). \bibinfo{title}{Surrogate
  gradient learning in spiking neural networks: Bringing the power of
  gradient-based optimization to spiking neural networks}.
\newblock \bibinfo{journal}{IEEE Signal Processing Magazine}
  \emph{\bibinfo{volume}{36}}, \bibinfo{pages}{51--63}.
  \DOIprefix\doi{https://doi.org/10.1109/MSP.2019.2931595}.
\bibitem[{Cramer et~al.(2022)Cramer, Billaudelle, Kanya, Leibfried, Grübl,
  Karasenko, Pehle, Schreiber, Stradmann, Weis, Schemmel and
  Zenke}]{cramer2022sg}
\bibinfo{author}{Cramer, B.}, \bibinfo{author}{Billaudelle, S.},
  \bibinfo{author}{Kanya, S.}, \bibinfo{author}{Leibfried, A.},
  \bibinfo{author}{Grübl, A.}, \bibinfo{author}{Karasenko, V.},
  \bibinfo{author}{Pehle, C.}, \bibinfo{author}{Schreiber, K.},
  \bibinfo{author}{Stradmann, Y.}, \bibinfo{author}{Weis, J.},
  \bibinfo{author}{Schemmel, J.}, and \bibinfo{author}{Zenke, F.}
  (\bibinfo{year}{2022}). \bibinfo{title}{Surrogate gradients for analog
  neuromorphic computing}.
\newblock \bibinfo{journal}{Proceedings of the National Academy of Sciences}
  \emph{\bibinfo{volume}{119}}, \bibinfo{pages}{e2109194119}.
  \DOIprefix\doi{https://doi.org/10.1073/pnas.2109194119}.
\bibitem[{Eshraghian et~al.(2021)Eshraghian, Ward, Neftci, Wang, Lenz, Dwivedi,
  Bennamoun, Jeong and Lu}]{eshraghian2021training}
\bibinfo{author}{Eshraghian, J.~K.}, \bibinfo{author}{Ward, M.},
  \bibinfo{author}{Neftci, E.}, \bibinfo{author}{Wang, X.},
  \bibinfo{author}{Lenz, G.}, \bibinfo{author}{Dwivedi, G.},
  \bibinfo{author}{Bennamoun, M.}, \bibinfo{author}{Jeong, D.~S.}, and
  \bibinfo{author}{Lu, W.~D.} (\bibinfo{year}{2021}). \bibinfo{title}{Training
  spiking neural networks using lessons from deep learning}.
\newblock \bibinfo{journal}{arXiv preprint arXiv:2109.12894}.
  \DOIprefix\doi{https://doi.org/10.48550/arXiv.2109.12894}.
\bibitem[{Bellec et~al.(2020)Bellec, Scherr, Subramoney, Hajek, Salaj,
  Legenstein and Maass}]{bellec2020solution}
\bibinfo{author}{Bellec, G.}, \bibinfo{author}{Scherr, F.},
  \bibinfo{author}{Subramoney, A.}, \bibinfo{author}{Hajek, E.},
  \bibinfo{author}{Salaj, D.}, \bibinfo{author}{Legenstein, R.}, and
  \bibinfo{author}{Maass, W.} (\bibinfo{year}{2020}). \bibinfo{title}{A
  solution to the learning dilemma for recurrent networks of spiking neurons}.
\newblock \bibinfo{journal}{Nature Communications} \emph{\bibinfo{volume}{11}},
  \bibinfo{pages}{1--15}.
  \DOIprefix\doi{https://doi.org/10.1038/s41467-020-17236-y}.
\bibitem[{Rumelhart et~al.(1986)Rumelhart, Hinton and
  Williams}]{rumelhart1986learning}
\bibinfo{author}{Rumelhart, D.~E.}, \bibinfo{author}{Hinton, G.~E.}, and
  \bibinfo{author}{Williams, R.~J.} (\bibinfo{year}{1986}).
  \bibinfo{title}{Learning representations by back-propagating errors}.
\newblock \bibinfo{journal}{Nature} \emph{\bibinfo{volume}{323}},
  \bibinfo{pages}{533--536}. \DOIprefix\doi{https://doi.org/10.1038/323533a0}.
\bibitem[{Zenke and Vogels(2021)}]{zenke2021remarkable}
\bibinfo{author}{Zenke, F.}, and \bibinfo{author}{Vogels, T.~P.}
  (\bibinfo{year}{2021}). \bibinfo{title}{The remarkable robustness of
  surrogate gradient learning for instilling complex function in spiking neural
  networks}.
\newblock \bibinfo{journal}{Neural Computation} \emph{\bibinfo{volume}{33}},
  \bibinfo{pages}{899--925}.
  \DOIprefix\doi{https://doi.org/10.1162/neco_a_01367}.
\bibitem[{Jang et~al.(2019)Jang, Simeone, Gardner and
  Gruning}]{jang2019introduction}
\bibinfo{author}{Jang, H.}, \bibinfo{author}{Simeone, O.},
  \bibinfo{author}{Gardner, B.}, and \bibinfo{author}{Gruning, A.}
  (\bibinfo{year}{2019}). \bibinfo{title}{An introduction to probabilistic
  spiking neural networks: Probabilistic models, learning rules, and
  applications}.
\newblock \bibinfo{journal}{IEEE Signal Processing Magazine}
  \emph{\bibinfo{volume}{36}}, \bibinfo{pages}{64--77}.
  \DOIprefix\doi{https://doi.org/10.1109/MSP.2019.2935234}.
\bibitem[{Dan and Poo(2004)}]{dan2004spike}
\bibinfo{author}{Dan, Y.}, and \bibinfo{author}{Poo, M.-m.}
  (\bibinfo{year}{2004}). \bibinfo{title}{Spike timing-dependent plasticity of
  neural circuits}.
\newblock \bibinfo{journal}{Neuron} \emph{\bibinfo{volume}{44}},
  \bibinfo{pages}{23--30}.
  \DOIprefix\doi{https://doi.org/10.1016/j.neuron.2004.09.007}.
\bibitem[{Froemke et~al.(2005)Froemke, Poo and Dan}]{froemke2005spike}
\bibinfo{author}{Froemke, R.~C.}, \bibinfo{author}{Poo, M.-m.}, and
  \bibinfo{author}{Dan, Y.} (\bibinfo{year}{2005}).
  \bibinfo{title}{Spike-timing-dependent synaptic plasticity depends on
  dendritic location}.
\newblock \bibinfo{journal}{Nature} \emph{\bibinfo{volume}{434}},
  \bibinfo{pages}{221--225}.
  \DOIprefix\doi{https://doi.org/10.1038/nature03366}.
\bibitem[{Guyonneau et~al.(2005)Guyonneau, VanRullen and
  Thorpe}]{guyonneau2005neurons}
\bibinfo{author}{Guyonneau, R.}, \bibinfo{author}{VanRullen, R.}, and
  \bibinfo{author}{Thorpe, S.~J.} (\bibinfo{year}{2005}).
  \bibinfo{title}{Neurons tune to the earliest spikes through stdp}.
\newblock \bibinfo{journal}{Neural Computation} \emph{\bibinfo{volume}{17}},
  \bibinfo{pages}{859--879}.
  \DOIprefix\doi{https://doi.org/10.1162/0899766053429390}.
\bibitem[{Maass(2014)}]{maass2014noise}
\bibinfo{author}{Maass, W.} (\bibinfo{year}{2014}). \bibinfo{title}{Noise as a
  resource for computation and learning in networks of spiking neurons}.
\newblock \bibinfo{journal}{Proceedings of the IEEE}
  \emph{\bibinfo{volume}{102}}, \bibinfo{pages}{860--880}.
  \DOIprefix\doi{https://doi.org/10.1109/JPROC.2014.2310593}.
\bibitem[{Burkitt(2006)}]{burkitt2006review}
\bibinfo{author}{Burkitt, A.~N.} (\bibinfo{year}{2006}). \bibinfo{title}{A
  review of the integrate-and-fire neuron model: I. homogeneous synaptic
  input}.
\newblock \bibinfo{journal}{Biological Cybernetics}
  \emph{\bibinfo{volume}{95}}, \bibinfo{pages}{1--19}.
  \DOIprefix\doi{https://doi.org/10.1007/s00422-006-0068-6}.
\bibitem[{Heckerman et~al.(1995)Heckerman, Geiger and
  Chickering}]{heckerman1995learning}
\bibinfo{author}{Heckerman, D.}, \bibinfo{author}{Geiger, D.}, and
  \bibinfo{author}{Chickering, D.~M.} (\bibinfo{year}{1995}).
  \bibinfo{title}{Learning bayesian networks: The combination of knowledge and
  statistical data}.
\newblock \bibinfo{journal}{Machine learning} \emph{\bibinfo{volume}{20}},
  \bibinfo{pages}{197--243}.
  \DOIprefix\doi{https://doi.org/10.1007/BF00994016}.
\bibitem[{Heckerman(1998)}]{heckerman1998tutorial}
\bibinfo{author}{Heckerman, D.}
\newblock \bibinfo{title}{A tutorial on learning with Bayesian networks}.
\newblock \bibinfo{publisher}{Springer} (\bibinfo{year}{1998}).
\newblock \DOIprefix\doi{https://doi.org/10.1007/978-94-011-5014-9_11}.
\bibitem[{Zenke and Neftci(2021)}]{zenke2021brain}
\bibinfo{author}{Zenke, F.}, and \bibinfo{author}{Neftci, E.~O.}
  (\bibinfo{year}{2021}). \bibinfo{title}{Brain-inspired learning on
  neuromorphic substrates}.
\newblock \bibinfo{journal}{Proceedings of the IEEE}
  \emph{\bibinfo{volume}{109}}, \bibinfo{pages}{935--950}.
  \DOIprefix\doi{https://doi.org/10.1109/JPROC.2020.3045625}.
\bibitem[{Wu et~al.(2022)Wu, Zhao, Zhu, Chen, Xu, Li, Song, Deng, Wang, Zheng
  et~al.}]{wu2022brain}
\bibinfo{author}{Wu, Y.}, \bibinfo{author}{Zhao, R.}, \bibinfo{author}{Zhu,
  J.}, \bibinfo{author}{Chen, F.}, \bibinfo{author}{Xu, M.},
  \bibinfo{author}{Li, G.}, \bibinfo{author}{Song, S.}, \bibinfo{author}{Deng,
  L.}, \bibinfo{author}{Wang, G.}, \bibinfo{author}{Zheng, H.} et~al.
  (\bibinfo{year}{2022}). \bibinfo{title}{Brain-inspired global-local learning
  incorporated with neuromorphic computing}.
\newblock \bibinfo{journal}{Nature Communications} \emph{\bibinfo{volume}{13}},
  \bibinfo{pages}{65}.
  \DOIprefix\doi{https://doi.org/10.1038/s41467-021-27653-2}.
\bibitem[{Fr{\'e}maux and Gerstner(2016)}]{fremaux2016neuromodulated}
\bibinfo{author}{Fr{\'e}maux, N.}, and \bibinfo{author}{Gerstner, W.}
  (\bibinfo{year}{2016}). \bibinfo{title}{Neuromodulated spike-timing-dependent
  plasticity, and theory of three-factor learning rules}.
\newblock \bibinfo{journal}{Frontiers in Neural Circuits}
  \emph{\bibinfo{volume}{9}}, \bibinfo{pages}{85}.
  \DOIprefix\doi{https://doi.org/10.3389/fncir.2015.00085}.
\bibitem[{Gerstner et~al.(2018)Gerstner, Lehmann, Liakoni, Corneil and
  Brea}]{gerstner2018eligibility}
\bibinfo{author}{Gerstner, W.}, \bibinfo{author}{Lehmann, M.},
  \bibinfo{author}{Liakoni, V.}, \bibinfo{author}{Corneil, D.}, and
  \bibinfo{author}{Brea, J.} (\bibinfo{year}{2018}).
  \bibinfo{title}{Eligibility traces and plasticity on behavioral time scales:
  experimental support of neohebbian three-factor learning rules}.
\newblock \bibinfo{journal}{Frontiers in Neural Circuits}
  \emph{\bibinfo{volume}{12}}, \bibinfo{pages}{53}.
  \DOIprefix\doi{https://doi.org/10.3389/fncir.2018.00053}.
\bibitem[{Hubara et~al.(2016)Hubara, Courbariaux, Soudry, El-Yaniv and
  Bengio}]{Hubara2016}
\bibinfo{author}{Hubara, I.}, \bibinfo{author}{Courbariaux, M.},
  \bibinfo{author}{Soudry, D.}, \bibinfo{author}{El-Yaniv, R.}, and
  \bibinfo{author}{Bengio, Y.}
\newblock \bibinfo{title}{{Binarized neural networks}} (\bibinfo{year}{2016}).
\newblock \DOIprefix\doi{https://dl.acm.org/doi/10.5555/3157382.3157557}.
\bibitem[{Tokui and Sato(2017)}]{tokui2017evaluating}
\bibinfo{author}{Tokui, S.}, and \bibinfo{author}{Sato, I.}
\newblock \bibinfo{title}{Evaluating the variance of likelihood-ratio gradient
  estimators}.
\newblock In: \emph{\bibinfo{booktitle}{International Conference on Machine
  Learning}}. \bibinfo{organization}{PMLR} (\bibinfo{year}{2017}):\unskip(
  \bibinfo{pages}{3414--3423}).
\newblock \DOIprefix\doi{https://dl.acm.org/doi/10.5555/3305890.3306034}.
\bibitem[{Hou et~al.(2017)Hou, Yao and Kwok}]{Hou2017}
\bibinfo{author}{Hou, L.}, \bibinfo{author}{Yao, Q.}, and
  \bibinfo{author}{Kwok, J.~T.}
\newblock \bibinfo{title}{{Loss-aware binarization of deep networks}}.
\newblock In: \emph{\bibinfo{booktitle}{International Conference on Learning
  Representations}} (\bibinfo{year}{2017}):\unskip( \bibinfo{pages}{1--11}).
\newblock \href{http://arxiv.org/abs/1611.01600}{\tt arXiv:1611.01600}.
\bibitem[{Yin et~al.(2019)Yin, Lyu, Zhang, Osher, Qi and Xin}]{Yin2019}
\bibinfo{author}{Yin, P.}, \bibinfo{author}{Lyu, J.}, \bibinfo{author}{Zhang,
  S.}, \bibinfo{author}{Osher, S.}, \bibinfo{author}{Qi, Y.}, and
  \bibinfo{author}{Xin, J.}
\newblock \bibinfo{title}{{Understanding straight-through estimator in training
  activation quantized neural nets}}.
\newblock In: \emph{\bibinfo{booktitle}{International Conference on Learning
  Representations}} (\bibinfo{year}{2019}):\unskip( \bibinfo{pages}{1--30}).
\newblock \DOIprefix\doi{https://doi.org/10.48550/arXiv.1903.05662}.
\bibitem[{Krizhevsky et~al.(2009)Krizhevsky, Hinton
  et~al.}]{krizhevsky2009learning}
\bibinfo{author}{Krizhevsky, A.}, \bibinfo{author}{Hinton, G.} et~al.
  (\bibinfo{year}{2009}). \bibinfo{title}{Learning multiple layers of features
  from tiny images}.
\bibitem[{Li et~al.(2017)Li, Liu, Ji, Li and Shi}]{li2017cifar10}
\bibinfo{author}{Li, H.}, \bibinfo{author}{Liu, H.}, \bibinfo{author}{Ji, X.},
  \bibinfo{author}{Li, G.}, and \bibinfo{author}{Shi, L.}
  (\bibinfo{year}{2017}). \bibinfo{title}{Cifar10-dvs: an event-stream dataset
  for object classification}.
\newblock \bibinfo{journal}{Frontiers in neuroscience}
  \emph{\bibinfo{volume}{11}}, \bibinfo{pages}{309}.
  \DOIprefix\doi{https://doi.org/10.3389/fnins.2017.00309}.
\bibitem[{Amir et~al.(2017)Amir, Taba, Berg, Melano, McKinstry, Di~Nolfo,
  Nayak, Andreopoulos, Garreau, Mendoza et~al.}]{amir2017low}
\bibinfo{author}{Amir, A.}, \bibinfo{author}{Taba, B.}, \bibinfo{author}{Berg,
  D.}, \bibinfo{author}{Melano, T.}, \bibinfo{author}{McKinstry, J.},
  \bibinfo{author}{Di~Nolfo, C.}, \bibinfo{author}{Nayak, T.},
  \bibinfo{author}{Andreopoulos, A.}, \bibinfo{author}{Garreau, G.},
  \bibinfo{author}{Mendoza, M.} et~al.
\newblock \bibinfo{title}{A low power, fully event-based gesture recognition
  system}.
\newblock In: \emph{\bibinfo{booktitle}{The IEEE / CVF Computer Vision and
  Pattern Recognition Conference}} (\bibinfo{year}{2017}):\unskip(
  \bibinfo{pages}{7243--7252}).
\newblock \DOIprefix\doi{https://doi.org/10.1109/CVPR.2017.781}.
\bibitem[{Basalyga and Salinas(2006)}]{basalyga2006response}
\bibinfo{author}{Basalyga, G.}, and \bibinfo{author}{Salinas, E.}
  (\bibinfo{year}{2006}). \bibinfo{title}{When response variability increases
  neural network robustness to synaptic noise}.
\newblock \bibinfo{journal}{Neural Computation} \emph{\bibinfo{volume}{18}},
  \bibinfo{pages}{1349--1379}.
  \DOIprefix\doi{https://doi.org/10.1162/neco.2006.18.6.1349}.
\bibitem[{McDonnell and Ward(2011)}]{mcdonnell2011benefits}
\bibinfo{author}{McDonnell, M.~D.}, and \bibinfo{author}{Ward, L.~M.}
  (\bibinfo{year}{2011}). \bibinfo{title}{The benefits of noise in neural
  systems: bridging theory and experiment}.
\newblock \bibinfo{journal}{Nature Reviews Neuroscience}
  \emph{\bibinfo{volume}{12}}, \bibinfo{pages}{415--425}.
  \DOIprefix\doi{https://doi.org/10.1038/nrn3061}.
\bibitem[{Gu et~al.(2021)Gu, Sng, Hu and Yu}]{gu2021eventdrop}
\bibinfo{author}{Gu, F.}, \bibinfo{author}{Sng, W.}, \bibinfo{author}{Hu, X.},
  and \bibinfo{author}{Yu, F.} (\bibinfo{year}{2021}).
  \bibinfo{title}{Eventdrop: Data augmentation for event-based learning}.
\newblock \bibinfo{journal}{arXiv}.
  \DOIprefix\doi{https://doi.org/10.48550/arXiv.2106.05836}.
\bibitem[{Mainen and Sejnowski(1995)}]{mainen1995reliability}
\bibinfo{author}{Mainen, Z.~F.}, and \bibinfo{author}{Sejnowski, T.~J.}
  (\bibinfo{year}{1995}). \bibinfo{title}{Reliability of spike timing in
  neocortical neurons}.
\newblock \bibinfo{journal}{Science} \emph{\bibinfo{volume}{268}},
  \bibinfo{pages}{1503--1506}.
  \DOIprefix\doi{https://doi.org/10.1126/science.7770778}.
\bibitem[{Tiesinga et~al.(2008)Tiesinga, Fellous and
  Sejnowski}]{tiesinga2008regulation}
\bibinfo{author}{Tiesinga, P.}, \bibinfo{author}{Fellous, J.-M.}, and
  \bibinfo{author}{Sejnowski, T.~J.} (\bibinfo{year}{2008}).
  \bibinfo{title}{Regulation of spike timing in visual cortical circuits}.
\newblock \bibinfo{journal}{Nature reviews neuroscience}
  \emph{\bibinfo{volume}{9}}, \bibinfo{pages}{97--107}.
  \DOIprefix\doi{https://doi.org/10.1038/nrn2315}.
\bibitem[{de~Ruyter~van Steveninck et~al.(1997)de~Ruyter~van Steveninck, Lewen,
  Strong, Koberle and Bialek}]{de1997reproducibility}
\bibinfo{author}{de~Ruyter~van Steveninck, R.~R.}, \bibinfo{author}{Lewen,
  G.~D.}, \bibinfo{author}{Strong, S.~P.}, \bibinfo{author}{Koberle, R.}, and
  \bibinfo{author}{Bialek, W.} (\bibinfo{year}{1997}).
  \bibinfo{title}{Reproducibility and variability in neural spike trains}.
\newblock \bibinfo{journal}{Science} \emph{\bibinfo{volume}{275}},
  \bibinfo{pages}{1805--1808}.
  \DOIprefix\doi{https://doi.org/10.1126/science.275.5307.1805}.
\bibitem[{Rolls and Deco(2010)}]{edmund2010}
\bibinfo{author}{Rolls, E.~T.}, and \bibinfo{author}{Deco, G.}
\newblock \bibinfo{title}{The Noisy Brain: Stochastic Dynamics as a Principle
  of Brain Function}.
\newblock \bibinfo{publisher}{Oxford University Press} (\bibinfo{year}{2010}).
\newblock \DOIprefix\doi{10.1093/acprof:oso/9780199587865.001.0001}.
\bibitem[{Onken et~al.(2016)Onken, Liu, Karunasekara, Delis, Gollisch and
  Panzeri}]{onken2016using}
\bibinfo{author}{Onken, A.}, \bibinfo{author}{Liu, J.~K.},
  \bibinfo{author}{Karunasekara, P. C.~R.}, \bibinfo{author}{Delis, I.},
  \bibinfo{author}{Gollisch, T.}, and \bibinfo{author}{Panzeri, S.}
  (\bibinfo{year}{2016}). \bibinfo{title}{Using matrix and tensor
  factorizations for the single-trial analysis of population spike trains}.
\newblock \bibinfo{journal}{PLoS Computational Biology}
  \emph{\bibinfo{volume}{12}}, \bibinfo{pages}{e1005189}.
  \DOIprefix\doi{https://doi.org/10.1371/journal.pcbi.1005189}.
\bibitem[{McIntosh et~al.(2016)McIntosh, Maheswaranathan, Nayebi, Ganguli and
  Baccus}]{mcintosh2016}
\bibinfo{author}{McIntosh, L.}, \bibinfo{author}{Maheswaranathan, N.},
  \bibinfo{author}{Nayebi, A.}, \bibinfo{author}{Ganguli, S.}, and
  \bibinfo{author}{Baccus, S.}
\newblock \bibinfo{title}{Deep learning models of the retinal response to
  natural scenes}.
\newblock In: \emph{\bibinfo{booktitle}{Advances in Neural Information
  Processing Systems (NeurIPS)}}
  (\bibinfo{year}{2016}):\unskip\DOIprefix\doi{https://dl.acm.org/doi/10.5555/3157096.3157249}.
\bibitem[{Zheng et~al.(2021{\natexlab{a}})Zheng, Jia, Yu, Liu and
  Huang}]{zheng2021unraveling}
\bibinfo{author}{Zheng, Y.}, \bibinfo{author}{Jia, S.}, \bibinfo{author}{Yu,
  Z.}, \bibinfo{author}{Liu, J.~K.}, and \bibinfo{author}{Huang, T.}
  (\bibinfo{year}{2021}{\natexlab{a}}). \bibinfo{title}{Unraveling neural
  coding of dynamic natural visual scenes via convolutional recurrent neural
  networks}.
\newblock \bibinfo{journal}{Patterns} \emph{\bibinfo{volume}{2}},
  \bibinfo{pages}{100350}.
  \DOIprefix\doi{https://doi.org/10.1016/j.patter.2021.100350}.
\bibitem[{Yamins et~al.(2014)Yamins, Hong, Cadieu, Solomon, Seibert and
  DiCarlo}]{yamins2014performance}
\bibinfo{author}{Yamins, D.~L.}, \bibinfo{author}{Hong, H.},
  \bibinfo{author}{Cadieu, C.~F.}, \bibinfo{author}{Solomon, E.~A.},
  \bibinfo{author}{Seibert, D.}, and \bibinfo{author}{DiCarlo, J.~J.}
  (\bibinfo{year}{2014}). \bibinfo{title}{Performance-optimized hierarchical
  models predict neural responses in higher visual cortex}.
\newblock \bibinfo{journal}{Proceedings of the National Academy of Sciences}
  \emph{\bibinfo{volume}{111}}, \bibinfo{pages}{8619--8624}.
  \DOIprefix\doi{https://doi.org/10.1073/pnas.1403112111}.
\bibitem[{Gerstner and Kistler(2002)}]{gerstner2002spiking}
\bibinfo{author}{Gerstner, W.}, and \bibinfo{author}{Kistler, W.~M.}
\newblock \bibinfo{title}{Spiking neuron models: Single neurons, populations,
  plasticity}.
\newblock \bibinfo{publisher}{Cambridge university press}
  (\bibinfo{year}{2002}).
\newblock \DOIprefix\doi{https://doi.org/10.1017/CBO9780511815706}.
\bibitem[{Vaswani et~al.(2017)Vaswani, Shazeer, Parmar, Uszkoreit, Jones,
  Gomez, Kaiser and Polosukhin}]{vaswani2017attention}
\bibinfo{author}{Vaswani, A.}, \bibinfo{author}{Shazeer, N.},
  \bibinfo{author}{Parmar, N.}, \bibinfo{author}{Uszkoreit, J.},
  \bibinfo{author}{Jones, L.}, \bibinfo{author}{Gomez, A.~N.},
  \bibinfo{author}{Kaiser, {\L}.}, and \bibinfo{author}{Polosukhin, I.}
\newblock \bibinfo{title}{Attention is all you need}.
\newblock In: \emph{\bibinfo{booktitle}{Advances in Neural Information
  Processing Systems (NeurIPS)}} vol.~\bibinfo{volume}{30}
  (\bibinfo{year}{2017}):\unskip.
\bibitem[{Zhang et~al.(2022)Zhang, Dong, Zhang, Ding, Heide, Yin and
  Yang}]{zhang2022spiking}
\bibinfo{author}{Zhang, J.}, \bibinfo{author}{Dong, B.},
  \bibinfo{author}{Zhang, H.}, \bibinfo{author}{Ding, J.},
  \bibinfo{author}{Heide, F.}, \bibinfo{author}{Yin, B.}, and
  \bibinfo{author}{Yang, X.}
\newblock \bibinfo{title}{Spiking transformers for event-based single object
  tracking}.
\newblock In: \emph{\bibinfo{booktitle}{The IEEE / CVF Computer Vision and
  Pattern Recognition Conference}} (\bibinfo{year}{2022}):\unskip(
  \bibinfo{pages}{8801--8810}).
\bibitem[{Indiveri and Liu(2015)}]{indiveri2015memory}
\bibinfo{author}{Indiveri, G.}, and \bibinfo{author}{Liu, S.-C.}
  (\bibinfo{year}{2015}). \bibinfo{title}{Memory and information processing in
  neuromorphic systems}.
\newblock \bibinfo{journal}{Proceedings of the IEEE}
  \emph{\bibinfo{volume}{103}}, \bibinfo{pages}{1379--1397}.
  \DOIprefix\doi{https://doi.org/10.1109/JPROC.2015.2444094}.
\bibitem[{Pei et~al.(2019)Pei, Deng, Song, Zhao, Zhang, Wu, Wang, Zou, Wu, He
  et~al.}]{pei2019towards}
\bibinfo{author}{Pei, J.}, \bibinfo{author}{Deng, L.}, \bibinfo{author}{Song,
  S.}, \bibinfo{author}{Zhao, M.}, \bibinfo{author}{Zhang, Y.},
  \bibinfo{author}{Wu, S.}, \bibinfo{author}{Wang, G.}, \bibinfo{author}{Zou,
  Z.}, \bibinfo{author}{Wu, Z.}, \bibinfo{author}{He, W.} et~al.
  (\bibinfo{year}{2019}). \bibinfo{title}{Towards artificial general
  intelligence with hybrid tianjic chip architecture}.
\newblock \bibinfo{journal}{Nature} \emph{\bibinfo{volume}{572}},
  \bibinfo{pages}{106--111}.
  \DOIprefix\doi{https://doi.org/10.1038/s41586-019-1424-8}.
\bibitem[{Davies et~al.(2021)Davies, Wild, Orchard, Sandamirskaya, Guerra,
  Joshi, Plank and Risbud}]{davies2021advancing}
\bibinfo{author}{Davies, M.}, \bibinfo{author}{Wild, A.},
  \bibinfo{author}{Orchard, G.}, \bibinfo{author}{Sandamirskaya, Y.},
  \bibinfo{author}{Guerra, G. A.~F.}, \bibinfo{author}{Joshi, P.},
  \bibinfo{author}{Plank, P.}, and \bibinfo{author}{Risbud, S.~R.}
  (\bibinfo{year}{2021}). \bibinfo{title}{Advancing neuromorphic computing with
  loihi: A survey of results and outlook}.
\newblock \bibinfo{journal}{Proceedings of the IEEE}
  \emph{\bibinfo{volume}{109}}, \bibinfo{pages}{911--934}.
  \DOIprefix\doi{https://doi.org/10.1109/JPROC.2021.3067593}.
\bibitem[{Benjamin et~al.(2014)Benjamin, Gao, McQuinn, Choudhary,
  Chandrasekaran, Bussat, Alvarez-Icaza, Arthur, Merolla and
  Boahen}]{benjamin2014neurogrid}
\bibinfo{author}{Benjamin, B.~V.}, \bibinfo{author}{Gao, P.},
  \bibinfo{author}{McQuinn, E.}, \bibinfo{author}{Choudhary, S.},
  \bibinfo{author}{Chandrasekaran, A.~R.}, \bibinfo{author}{Bussat, J.-M.},
  \bibinfo{author}{Alvarez-Icaza, R.}, \bibinfo{author}{Arthur, J.~V.},
  \bibinfo{author}{Merolla, P.~A.}, and \bibinfo{author}{Boahen, K.}
  (\bibinfo{year}{2014}). \bibinfo{title}{Neurogrid: A mixed-analog-digital
  multichip system for large-scale neural simulations}.
\newblock \bibinfo{journal}{Proceedings of the IEEE}
  \emph{\bibinfo{volume}{102}}, \bibinfo{pages}{699--716}.
  \DOIprefix\doi{https://doi.org/10.1109/JPROC.2014.2313565}.
\bibitem[{DeBole et~al.(2019)DeBole, Taba, Amir, Akopyan, Andreopoulos, Risk,
  Kusnitz, Otero, Nayak, Appuswamy et~al.}]{debole2019truenorth}
\bibinfo{author}{DeBole, M.~V.}, \bibinfo{author}{Taba, B.},
  \bibinfo{author}{Amir, A.}, \bibinfo{author}{Akopyan, F.},
  \bibinfo{author}{Andreopoulos, A.}, \bibinfo{author}{Risk, W.~P.},
  \bibinfo{author}{Kusnitz, J.}, \bibinfo{author}{Otero, C.~O.},
  \bibinfo{author}{Nayak, T.~K.}, \bibinfo{author}{Appuswamy, R.} et~al.
  (\bibinfo{year}{2019}). \bibinfo{title}{Truenorth: Accelerating from zero to
  64 million neurons in 10 years}.
\newblock \bibinfo{journal}{Computer} \emph{\bibinfo{volume}{52}},
  \bibinfo{pages}{20--29}.
  \DOIprefix\doi{https://doi.org/10.1109/MC.2019.2903009}.
\bibitem[{Davies et~al.(2018)Davies, Srinivasa, Lin, Chinya, Cao, Choday,
  Dimou, Joshi, Imam, Jain, Liao, Lin, Lines, Liu, Mathaikutty, McCoy, Paul,
  Tse, Venkataramanan, Weng, Wild, Yang and Wang}]{loihi2018}
\bibinfo{author}{Davies, M.}, \bibinfo{author}{Srinivasa, N.},
  \bibinfo{author}{Lin, T.-H.}, \bibinfo{author}{Chinya, G.},
  \bibinfo{author}{Cao, Y.}, \bibinfo{author}{Choday, S.~H.},
  \bibinfo{author}{Dimou, G.}, \bibinfo{author}{Joshi, P.},
  \bibinfo{author}{Imam, N.}, \bibinfo{author}{Jain, S.},
  \bibinfo{author}{Liao, Y.}, \bibinfo{author}{Lin, C.-K.},
  \bibinfo{author}{Lines, A.}, \bibinfo{author}{Liu, R.},
  \bibinfo{author}{Mathaikutty, D.}, \bibinfo{author}{McCoy, S.},
  \bibinfo{author}{Paul, A.}, \bibinfo{author}{Tse, J.},
  \bibinfo{author}{Venkataramanan, G.}, \bibinfo{author}{Weng, Y.-H.},
  \bibinfo{author}{Wild, A.}, \bibinfo{author}{Yang, Y.}, and
  \bibinfo{author}{Wang, H.} (\bibinfo{year}{2018}). \bibinfo{title}{Loihi: A
  neuromorphic manycore processor with on-chip learning}.
\newblock \bibinfo{journal}{IEEE Micro} \emph{\bibinfo{volume}{38}},
  \bibinfo{pages}{82--99}.
  \DOIprefix\doi{https://doi.org/10.1109/MM.2018.112130359}.
\bibitem[{Qiao et~al.(2015)Qiao, Mostafa, Corradi, Osswald, Stefanini,
  Sumislawska and Indiveri}]{qiao2015reconfigurable}
\bibinfo{author}{Qiao, N.}, \bibinfo{author}{Mostafa, H.},
  \bibinfo{author}{Corradi, F.}, \bibinfo{author}{Osswald, M.},
  \bibinfo{author}{Stefanini, F.}, \bibinfo{author}{Sumislawska, D.}, and
  \bibinfo{author}{Indiveri, G.} (\bibinfo{year}{2015}). \bibinfo{title}{A
  reconfigurable on-line learning spiking neuromorphic processor comprising 256
  neurons and 128k synapses}.
\newblock \bibinfo{journal}{Frontiers in neuroscience}
  \emph{\bibinfo{volume}{9}}, \bibinfo{pages}{141}.
  \DOIprefix\doi{https://doi.org/10.3389/fnins.2015.00141}.
\bibitem[{Hazan and Ezra~Tsur(2022)}]{hazan2022neuromorphic}
\bibinfo{author}{Hazan, A.}, and \bibinfo{author}{Ezra~Tsur, E.}
  (\bibinfo{year}{2022}). \bibinfo{title}{Neuromorphic neural engineering
  framework-inspired online continuous learning with analog circuitry}.
\newblock \bibinfo{journal}{Applied Sciences} \emph{\bibinfo{volume}{12}},
  \bibinfo{pages}{4528}. \DOIprefix\doi{https://doi.org/10.3390/app12094528}.
\bibitem[{White et~al.(2000)White, Rubinstein and Kay}]{white2000channel}
\bibinfo{author}{White, J.~A.}, \bibinfo{author}{Rubinstein, J.~T.}, and
  \bibinfo{author}{Kay, A.~R.} (\bibinfo{year}{2000}). \bibinfo{title}{Channel
  noise in neurons}.
\newblock \bibinfo{journal}{Trends in Neurosciences}
  \emph{\bibinfo{volume}{23}}, \bibinfo{pages}{131--137}.
  \DOIprefix\doi{https://doi.org/10.1016/S0166-2236(99)01521-0}.
\bibitem[{Grossman et~al.(2019)Grossman, Yeagle, Harel, Espinal, Harpaz, Noy,
  M{\'e}gevand, Groppe, Mehta and Malach}]{grossman2019noisy}
\bibinfo{author}{Grossman, S.}, \bibinfo{author}{Yeagle, E.~M.},
  \bibinfo{author}{Harel, M.}, \bibinfo{author}{Espinal, E.},
  \bibinfo{author}{Harpaz, R.}, \bibinfo{author}{Noy, N.},
  \bibinfo{author}{M{\'e}gevand, P.}, \bibinfo{author}{Groppe, D.~M.},
  \bibinfo{author}{Mehta, A.~D.}, and \bibinfo{author}{Malach, R.}
  (\bibinfo{year}{2019}). \bibinfo{title}{The noisy brain: power of
  resting-state fluctuations predicts individual recognition performance}.
\newblock \bibinfo{journal}{Cell reports} \emph{\bibinfo{volume}{29}},
  \bibinfo{pages}{3775--3784}.
  \DOIprefix\doi{https://doi.org/10.1016/j.celrep.2019.11.081}.
\bibitem[{Masland(2004)}]{masland2004neuronal}
\bibinfo{author}{Masland, R.~H.} (\bibinfo{year}{2004}).
  \bibinfo{title}{Neuronal cell types}.
\newblock \bibinfo{journal}{Current Biology} \emph{\bibinfo{volume}{14}},
  \bibinfo{pages}{R497--R500}. \DOIprefix\doi{10.1016/j.cub.2004.06.035}.
\bibitem[{Klindt et~al.(2017)Klindt, Ecker, Euler and
  Bethge}]{klindt2017neural}
\bibinfo{author}{Klindt, D.}, \bibinfo{author}{Ecker, A.~S.},
  \bibinfo{author}{Euler, T.}, and \bibinfo{author}{Bethge, M.}
\newblock \bibinfo{title}{Neural system identification for large populations
  separating “what” and “where”}.
\newblock In: \emph{\bibinfo{booktitle}{Advances in Neural Information
  Processing Systems (NeurIPS)}} vol.~\bibinfo{volume}{30}
  (\bibinfo{year}{2017}):\unskip( \bibinfo{pages}{3509--3519}).
\newblock \DOIprefix\doi{https://dl.acm.org/doi/10.5555/3294996.3295109}.
\bibitem[{Zhuang et~al.(2021)Zhuang, Yan, Nayebi, Schrimpf, Frank, DiCarlo and
  Yamins}]{zhuang2021unsupervised}
\bibinfo{author}{Zhuang, C.}, \bibinfo{author}{Yan, S.},
  \bibinfo{author}{Nayebi, A.}, \bibinfo{author}{Schrimpf, M.},
  \bibinfo{author}{Frank, M.~C.}, \bibinfo{author}{DiCarlo, J.~J.}, and
  \bibinfo{author}{Yamins, D.~L.} (\bibinfo{year}{2021}).
  \bibinfo{title}{Unsupervised neural network models of the ventral visual
  stream}.
\newblock \bibinfo{journal}{Proceedings of the National Academy of Sciences}
  \emph{\bibinfo{volume}{118}}, \bibinfo{pages}{e2014196118}.
  \DOIprefix\doi{https://doi.org/10.1073/pnas.2014196118}.
\bibitem[{Cadena et~al.(2019)Cadena, Denfield, Walker, Gatys, Tolias, Bethge
  and Ecker}]{cadena2019deep}
\bibinfo{author}{Cadena, S.~A.}, \bibinfo{author}{Denfield, G.~H.},
  \bibinfo{author}{Walker, E.~Y.}, \bibinfo{author}{Gatys, L.~A.},
  \bibinfo{author}{Tolias, A.~S.}, \bibinfo{author}{Bethge, M.}, and
  \bibinfo{author}{Ecker, A.~S.} (\bibinfo{year}{2019}). \bibinfo{title}{Deep
  convolutional models improve predictions of macaque v1 responses to natural
  images}.
\newblock \bibinfo{journal}{PLoS Computational Biology}
  \emph{\bibinfo{volume}{15}}, \bibinfo{pages}{e1006897}.
  \DOIprefix\doi{https://doi.org/10.1371/journal.pcbi.1006897}.
\bibitem[{Ratan~Murty et~al.(2021)Ratan~Murty, Bashivan, Abate, DiCarlo and
  Kanwisher}]{ratan2021computational}
\bibinfo{author}{Ratan~Murty, N.~A.}, \bibinfo{author}{Bashivan, P.},
  \bibinfo{author}{Abate, A.}, \bibinfo{author}{DiCarlo, J.~J.}, and
  \bibinfo{author}{Kanwisher, N.} (\bibinfo{year}{2021}).
  \bibinfo{title}{Computational models of category-selective brain regions
  enable high-throughput tests of selectivity}.
\newblock \bibinfo{journal}{Nature Communications} \emph{\bibinfo{volume}{12}},
  \bibinfo{pages}{5540}.
  \DOIprefix\doi{https://doi.org/10.1038/s41467-021-25409-6}.
\bibitem[{Ma(2023)}]{ma_2023_7986394}
\bibinfo{author}{Ma, G.}
\newblock \bibinfo{title}{genema/noisy-spiking-neuron-nets: Ver 0.0.1}
  (\bibinfo{year}{2023}).
\newblock \DOIprefix\doi{10.5281/zenodo.7986394}.
\bibitem[{Tal and Schwartz(1997)}]{tal1997computing}
\bibinfo{author}{Tal, D.}, and \bibinfo{author}{Schwartz, E.~L.}
  (\bibinfo{year}{1997}). \bibinfo{title}{Computing with the leaky
  integrate-and-fire neuron: logarithmic computation and multiplication}.
\newblock \bibinfo{journal}{Neural computation} \emph{\bibinfo{volume}{9}},
  \bibinfo{pages}{305--318}.
  \DOIprefix\doi{https://doi.org/10.1162/neco.1997.9.2.305}.
\bibitem[{Brunel and Van~Rossum(2007)}]{brunel2007lapicque}
\bibinfo{author}{Brunel, N.}, and \bibinfo{author}{Van~Rossum, M.~C.}
  (\bibinfo{year}{2007}). \bibinfo{title}{Lapicque’s 1907 paper: from frogs
  to integrate-and-fire}.
\newblock \bibinfo{journal}{Biological Cybernetics}
  \emph{\bibinfo{volume}{97}}, \bibinfo{pages}{337--339}.
  \DOIprefix\doi{https://doi.org/10.1007/s00422-007-0190-0}.
\bibitem[{Xiao et~al.(2022)Xiao, Meng, Zhang, He and Lin}]{xiao2022online}
\bibinfo{author}{Xiao, M.}, \bibinfo{author}{Meng, Q.}, \bibinfo{author}{Zhang,
  Z.}, \bibinfo{author}{He, D.}, and \bibinfo{author}{Lin, Z.}
\newblock \bibinfo{title}{Online training through time for spiking neural
  networks}.
\newblock In: \emph{\bibinfo{booktitle}{Advances in Neural Information
  Processing Systems (NeurIPS)}}
  (\bibinfo{year}{2022}):\unskip\DOIprefix\doi{http://dx.doi.org/10.48550/arXiv.2210.04195}.
\bibitem[{Van~Kampen(1992)}]{van1992stochastic}
\bibinfo{author}{Van~Kampen, N.~G.}
\newblock \bibinfo{title}{Stochastic processes in physics and chemistry}
  vol.~\bibinfo{volume}{1}.
\newblock \bibinfo{publisher}{Elsevier} (\bibinfo{year}{1992}).
\newblock \DOIprefix\doi{Stochastic processes in physics and chemistry}.
\bibitem[{Kloeden and Platen(1992)}]{kloeden1992stochastic}
\bibinfo{author}{Kloeden, P.~E.}, and \bibinfo{author}{Platen, E.}
\newblock \bibinfo{title}{Stochastic differential equations}.
\newblock In: \emph{\bibinfo{booktitle}{Numerical solution of stochastic
  differential equations}} ( \bibinfo{pages}{103--160}).
  \bibinfo{publisher}{Springer} (\bibinfo{year}{1992}):\unskip(
  \bibinfo{pages}{103--160}).
\newblock \DOIprefix\doi{https://doi.org/10.1007/978-3-662-12616-5_4}.
\bibitem[{Barndorff-Nielsen and Shephard(2001)}]{barndorff2001non}
\bibinfo{author}{Barndorff-Nielsen, O.~E.}, and \bibinfo{author}{Shephard, N.}
  (\bibinfo{year}{2001}). \bibinfo{title}{Non-gaussian ornstein-uhlenbeck-based
  models and some of their uses in financial economics}.
\newblock \bibinfo{journal}{Journal of the Royal Statistical Society: Series B
  (Statistical Methodology)} \emph{\bibinfo{volume}{63}},
  \bibinfo{pages}{167--241}.
  \DOIprefix\doi{https://doi.org/10.1111/1467-9868.00282}.
\bibitem[{Patel and Kosko(2008)}]{patel2008stochastic}
\bibinfo{author}{Patel, A.}, and \bibinfo{author}{Kosko, B.}
  (\bibinfo{year}{2008}). \bibinfo{title}{Stochastic resonance in continuous
  and spiking neuron models with levy noise}.
\newblock \bibinfo{journal}{IEEE Transactions on Neural Networks}
  \emph{\bibinfo{volume}{19}}, \bibinfo{pages}{1993--2008}.
  \DOIprefix\doi{https://doi.org/10.1109/TNN.2008.2005610}.
\bibitem[{Plesser and Gerstner(2000{\natexlab{b}})}]{plesser2000escape2}
\bibinfo{author}{Plesser, H.~E.}, and \bibinfo{author}{Gerstner, W.}
  (\bibinfo{year}{2000}{\natexlab{b}}). \bibinfo{title}{Escape rate models for
  noisy integrate-and-free neurons}.
\newblock \bibinfo{journal}{Neurocomputing} \emph{\bibinfo{volume}{32}},
  \bibinfo{pages}{219--224}.
  \DOIprefix\doi{https://doi.org/10.1016/S0925-2312(00)00167-3}.
\bibitem[{Jolivet et~al.(2006)Jolivet, Rauch, L{\"u}scher and
  Gerstner}]{jolivet2006predicting}
\bibinfo{author}{Jolivet, R.}, \bibinfo{author}{Rauch, A.},
  \bibinfo{author}{L{\"u}scher, H.-R.}, and \bibinfo{author}{Gerstner, W.}
  (\bibinfo{year}{2006}). \bibinfo{title}{Predicting spike timing of
  neocortical pyramidal neurons by simple threshold models}.
\newblock \bibinfo{journal}{Journal of computational neuroscience}
  \emph{\bibinfo{volume}{21}}, \bibinfo{pages}{35--49}.
  \DOIprefix\doi{https://doi.org/10.1007/s10827-006-7074-5}.
\bibitem[{Burt~Jr and Garman(1971)}]{burt1971conditional}
\bibinfo{author}{Burt~Jr, J.~M.}, and \bibinfo{author}{Garman, M.~B.}
  (\bibinfo{year}{1971}). \bibinfo{title}{Conditional monte carlo: A simulation
  technique for stochastic network analysis}.
\newblock \bibinfo{journal}{Management Science} \emph{\bibinfo{volume}{18}},
  \bibinfo{pages}{207--217}.
  \DOIprefix\doi{https://doi.org/10.1287/mnsc.18.3.207}.
\bibitem[{Titsias et~al.(2015)Titsias, L{\'a}zaro-Gredilla
  et~al.}]{aueb2015local}
\bibinfo{author}{Titsias, M.~K.}, \bibinfo{author}{L{\'a}zaro-Gredilla, M.}
  et~al.
\newblock \bibinfo{title}{Local expectation gradients for black box variational
  inference}.
\newblock In: \emph{\bibinfo{booktitle}{Advances in Neural Information
  Processing Systems (NeurIPS)}} (\bibinfo{year}{2015}):\unskip(
  \bibinfo{pages}{2638--2646}).
\newblock \DOIprefix\doi{https://dl.acm.org/doi/abs/10.5555/2969442.2969534}.
\bibitem[{Fiete and Seung(2006)}]{fiete2006gradient}
\bibinfo{author}{Fiete, I.~R.}, and \bibinfo{author}{Seung, H.~S.}
  (\bibinfo{year}{2006}). \bibinfo{title}{Gradient learning in spiking neural
  networks by dynamic perturbation of conductances}.
\newblock \bibinfo{journal}{Physical Review Letters}
  \emph{\bibinfo{volume}{97}}, \bibinfo{pages}{048104}.
  \DOIprefix\doi{https://doi.org/10.1103/PhysRevLett.97.048104}.
\bibitem[{Shekhovtsov et~al.(2020)Shekhovtsov, Yanush and
  Flach}]{shekhovtsov2020path}
\bibinfo{author}{Shekhovtsov, A.}, \bibinfo{author}{Yanush, V.}, and
  \bibinfo{author}{Flach, B.}
\newblock \bibinfo{title}{Path sample-analytic gradient estimators for
  stochastic binary networks}.
\newblock In: \emph{\bibinfo{booktitle}{Advances in Neural Information
  Processing Systems (NeurIPS)}} (\bibinfo{year}{2020}):\unskip(
  \bibinfo{pages}{12884--12894}).
\newblock \DOIprefix\doi{https://doi.org/10.48550/arXiv.2006.03143}.
\bibitem[{Shrestha and Orchard(2018)}]{shrestha2018slayer}
\bibinfo{author}{Shrestha, S.~B.}, and \bibinfo{author}{Orchard, G.}
\newblock \bibinfo{title}{Slayer: Spike layer error reassignment in time}.
\newblock In: \emph{\bibinfo{booktitle}{Advances in Neural Information
  Processing Systems (NeurIPS)}} vol.~\bibinfo{volume}{31}
  (\bibinfo{year}{2018}):\unskip\DOIprefix\doi{https://doi.org/10.48550/arXiv.1810.08646}.
\bibitem[{Mao(2007)}]{mao2007stochastic}
\bibinfo{author}{Mao, X.}
\newblock \bibinfo{title}{Stochastic differential equations and applications}.
\newblock \bibinfo{publisher}{Elsevier} (\bibinfo{year}{2007}).
\newblock \DOIprefix\doi{https://doi.org/10.1533/9780857099402.47}.
\bibitem[{Cubuk et~al.(2018)Cubuk, Zoph, Mane, Vasudevan and
  Le}]{cubuk2018autoaugment}
\bibinfo{author}{Cubuk, E.~D.}, \bibinfo{author}{Zoph, B.},
  \bibinfo{author}{Mane, D.}, \bibinfo{author}{Vasudevan, V.}, and
  \bibinfo{author}{Le, Q.~V.}
\newblock \bibinfo{title}{Autoaugment: Learning augmentation policies from
  data}.
\newblock In: \emph{\bibinfo{booktitle}{The IEEE / CVF Computer Vision and
  Pattern Recognition Conference}}
  (\bibinfo{year}{2018}):\unskip\DOIprefix\doi{https://doi.org/10.48550/arXiv.1805.09501}.
\bibitem[{Samadzadeh et~al.(2020)Samadzadeh, Far, Javadi, Nickabadi and
  Chehreghani}]{samadzadeh2020convolutional}
\bibinfo{author}{Samadzadeh, A.}, \bibinfo{author}{Far, F. S.~T.},
  \bibinfo{author}{Javadi, A.}, \bibinfo{author}{Nickabadi, A.}, and
  \bibinfo{author}{Chehreghani, M.~H.} (\bibinfo{year}{2020}).
  \bibinfo{title}{Convolutional spiking neural networks for spatio-temporal
  feature extraction}.
\newblock \bibinfo{journal}{arXiv}. \DOIprefix\doi{10.1007/s11063-023-11247-8}.
\bibitem[{Kingma and Ba(2014)}]{kingma2014adam}
\bibinfo{author}{Kingma, D.~P.}, and \bibinfo{author}{Ba, J.}
  (\bibinfo{year}{2014}). \bibinfo{title}{Adam: A method for stochastic
  optimization}.
\newblock \bibinfo{journal}{arXiv}.
  \DOIprefix\doi{https://doi.org/10.48550/arXiv.1412.6980}.
\bibitem[{Loshchilov and Hutter(2016)}]{loshchilov2016sgdr}
\bibinfo{author}{Loshchilov, I.}, and \bibinfo{author}{Hutter, F.}
  (\bibinfo{year}{2016}). \bibinfo{title}{Sgdr: Stochastic gradient descent
  with warm restarts}.
\newblock \bibinfo{journal}{arXiv}.
  \DOIprefix\doi{https://doi.org/10.48550/arXiv.1608.03983}.
\bibitem[{Zheng et~al.(2021{\natexlab{b}})Zheng, Wu, Deng, Hu and
  Li}]{zheng2021going}
\bibinfo{author}{Zheng, H.}, \bibinfo{author}{Wu, Y.}, \bibinfo{author}{Deng,
  L.}, \bibinfo{author}{Hu, Y.}, and \bibinfo{author}{Li, G.}
\newblock \bibinfo{title}{Going deeper with directly-trained larger spiking
  neural networks}.
\newblock In: \emph{\bibinfo{booktitle}{Proceedings of the AAAI conference on
  artificial intelligence}} vol.~\bibinfo{volume}{35}
  (\bibinfo{year}{2021}{\natexlab{b}}):\unskip( \bibinfo{pages}{11062--11070}).
\newblock \DOIprefix\doi{http://dx.doi.org/10.1609/aaai.v35i12.17320}.
\bibitem[{Gu et~al.(2019)Gu, Xiao, Pan and Tang}]{Gu2019}
\bibinfo{author}{Gu, P.}, \bibinfo{author}{Xiao, R.}, \bibinfo{author}{Pan,
  G.}, and \bibinfo{author}{Tang, H.}
\newblock \bibinfo{title}{Stca: Spatio-temporal credit assignment with delayed
  feedback in deep spiking neural networks}.
\newblock In: \emph{\bibinfo{booktitle}{International Joint Conferences on
  Artifical Intelligence}} (\bibinfo{year}{2019}):\unskip(
  \bibinfo{pages}{1366--1372}).
\newblock \DOIprefix\doi{https://doi.org/10.24963/ijcai.2019/189}.
\bibitem[{Wu et~al.(2021)Wu, Zhang, Lin, Li, Wang and Tang}]{wu2021liaf}
\bibinfo{author}{Wu, Z.}, \bibinfo{author}{Zhang, H.}, \bibinfo{author}{Lin,
  Y.}, \bibinfo{author}{Li, G.}, \bibinfo{author}{Wang, M.}, and
  \bibinfo{author}{Tang, Y.} (\bibinfo{year}{2021}). \bibinfo{title}{Liaf-net:
  Leaky integrate and analog fire network for lightweight and efficient
  spatiotemporal information processing}.
\newblock \bibinfo{journal}{IEEE Transactions on Neural Networks and Learning
  Systems}. \DOIprefix\doi{https://doi.org/10.1109/TNNLS.2021.3073016}.
\bibitem[{Liu et~al.(2021)Liu, Yang, Zhu, Lei, Cai, Wang, Huan and
  Lin}]{liu2021neuronal}
\bibinfo{author}{Liu, J.}, \bibinfo{author}{Yang, X.}, \bibinfo{author}{Zhu,
  Y.}, \bibinfo{author}{Lei, Y.}, \bibinfo{author}{Cai, J.},
  \bibinfo{author}{Wang, M.}, \bibinfo{author}{Huan, Z.}, and
  \bibinfo{author}{Lin, X.} (\bibinfo{year}{2021}). \bibinfo{title}{How
  neuronal noises influence the spiking neural networks’s cognitive learning
  process: A preliminary study}.
\newblock \bibinfo{journal}{Brain Sciences} \emph{\bibinfo{volume}{11}},
  \bibinfo{pages}{153}.
  \DOIprefix\doi{https://doi.org/10.3390/brainsci11020153}.
\bibitem[{Lezcano-Casado(2019)}]{lezcano2019trivializations}
\bibinfo{author}{Lezcano-Casado, M.}
\newblock \bibinfo{title}{Trivializations for gradient-based optimization on
  manifolds}.
\newblock In: \emph{\bibinfo{booktitle}{Advances in Neural Information
  Processing Systems (NeurIPS)}} (\bibinfo{year}{2019}):\unskip(
  \bibinfo{pages}{9154--9164}).
\newblock \DOIprefix\doi{https://doi.org/10.48550/arXiv.1909.09501}.
\bibitem[{Goodfellow et~al.(2015)Goodfellow, Shlens and
  Szegedy}]{goodfellow2014explaining}
\bibinfo{author}{Goodfellow, I.~J.}, \bibinfo{author}{Shlens, J.}, and
  \bibinfo{author}{Szegedy, C.}
\newblock \bibinfo{title}{Explaining and harnessing adversarial examples}.
\newblock In: \emph{\bibinfo{booktitle}{International Conference on Learning
  Representations}}
  (\bibinfo{year}{2015}):\unskip\href{http://arxiv.org/abs/1412.6572}{\tt
  arXiv:1412.6572}.
\bibitem[{Fano(1947)}]{fano1947ionization}
\bibinfo{author}{Fano, U.} (\bibinfo{year}{1947}). \bibinfo{title}{Ionization
  yield of radiations. ii. the fluctuations of the number of ions}.
\newblock \bibinfo{journal}{Physical Review} \emph{\bibinfo{volume}{72}},
  \bibinfo{pages}{26}. \DOIprefix\doi{https://doi.org/10.1103/PhysRev.72.26}.
\bibitem[{Park et~al.(2013)Park, Seth, Paiva, Li and Principe}]{park2013kernel}
\bibinfo{author}{Park, I.~M.}, \bibinfo{author}{Seth, S.},
  \bibinfo{author}{Paiva, A.~R.}, \bibinfo{author}{Li, L.}, and
  \bibinfo{author}{Principe, J.~C.} (\bibinfo{year}{2013}).
  \bibinfo{title}{Kernel methods on spike train space for neuroscience: a
  tutorial}.
\newblock \bibinfo{journal}{IEEE Signal Processing Magazine}
  \emph{\bibinfo{volume}{30}}, \bibinfo{pages}{149--160}.
  \DOIprefix\doi{https://doi.org/10.1109/MSP.2013.2251072}.
\bibitem[{Arribas et~al.(2020)Arribas, Zhao and Park}]{arribas2020rescuing}
\bibinfo{author}{Arribas, D.}, \bibinfo{author}{Zhao, Y.}, and
  \bibinfo{author}{Park, I.~M.}
\newblock \bibinfo{title}{Rescuing neural spike train models from bad mle}.
\newblock In: \emph{\bibinfo{booktitle}{Advances in Neural Information
  Processing Systems (NeurIPS)}} vol.~\bibinfo{volume}{33}
  (\bibinfo{year}{2020}):\unskip( \bibinfo{pages}{2293--2303}).
\newblock \DOIprefix\doi{https://dl.acm.org/doi/abs/10.5555/3495724.3495917}.
\bibitem[{Zenke and Ganguli(2018)}]{zenke2018superspike}
\bibinfo{author}{Zenke, F.}, and \bibinfo{author}{Ganguli, S.}
  (\bibinfo{year}{2018}). \bibinfo{title}{Superspike: Supervised learning in
  multilayer spiking neural networks}.
\newblock \bibinfo{journal}{Neural computation} \emph{\bibinfo{volume}{30}},
  \bibinfo{pages}{1514--1541}.
  \DOIprefix\doi{https://doi.org/10.1162/neco_a_01086}.

\end{thebibliography}

\newpage
\section*{Figure titles and legends}

\clearpage 

\section*{Tables, table titles, and table legends}

\begin{table}[htb]
    \caption{
    	{\bf Evaluation results on CIFAR-10, CIFAR-100, DVS-CIFAR and DVS-Gesture datasets.}  Data are represented as mean$\pm$SD. 
    }
    \includegraphics[width=1 \textwidth]{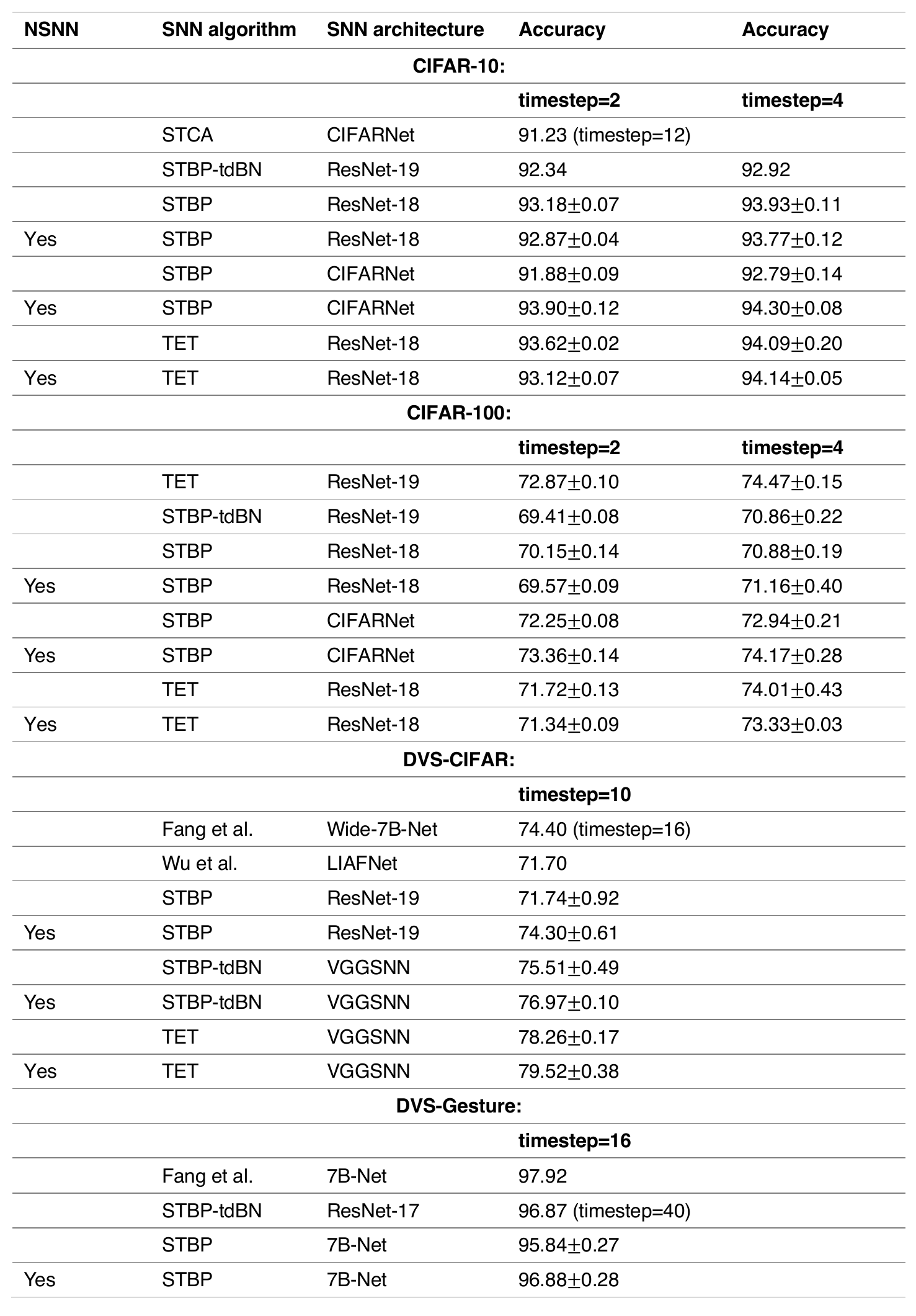}
    
    \label{tab:1}
\end{table}

\begin{table}[htb]
    \caption{
    	{\bf Evaluation results under EventDrop input-level perturbation on DVS-CIFAR data. } 
    	A larger $\rho$ denotes stronger perturbation. Data are represented as mean$\pm$SD. 
    }
    \includegraphics[width=1 \textwidth]{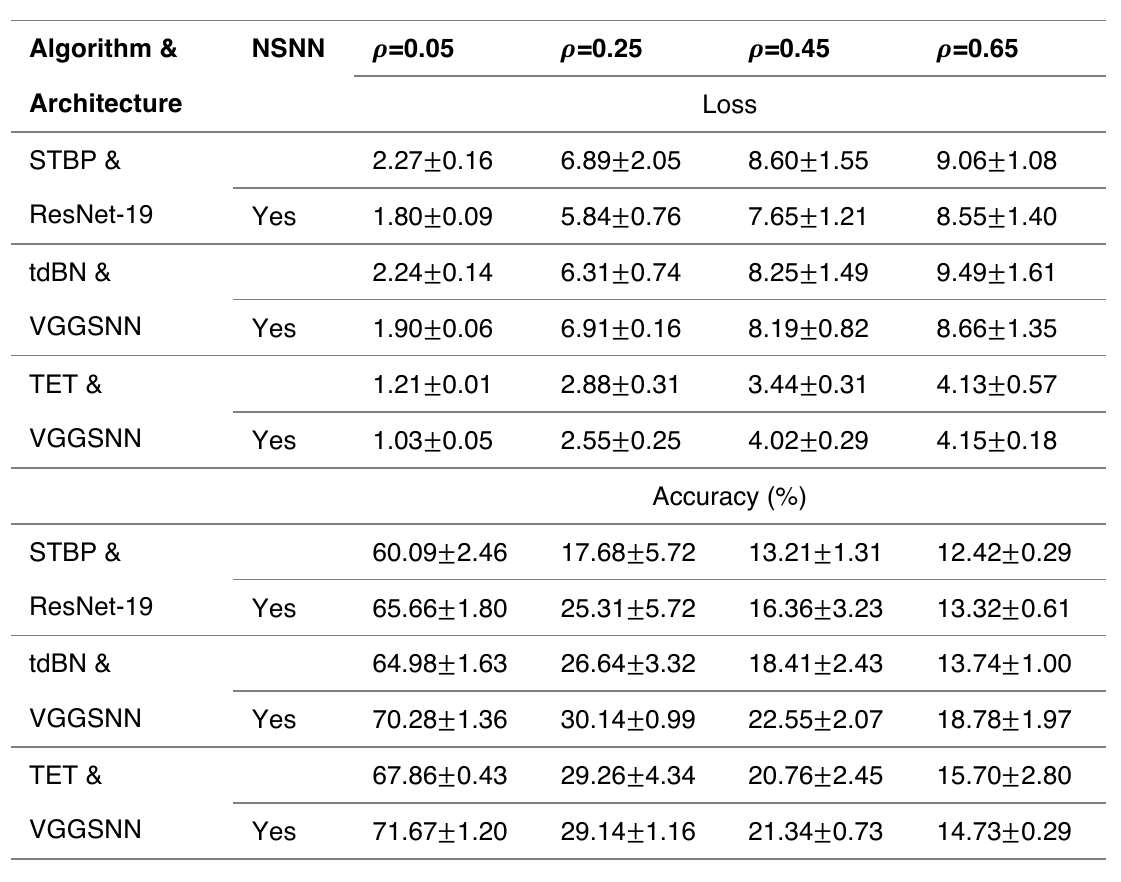}
    \label{tab:5}
\end{table}

\begin{table}[htb]
    \caption{
    	{\bf Evaluation results under spike state-level (firing state of spiking neurons) perturbation on DVS-CIFAR data. } 
    A larger $\beta$ denotes stronger perturbation. Data are represented as mean$\pm$SD.
    }
    \includegraphics[width=1 \textwidth]{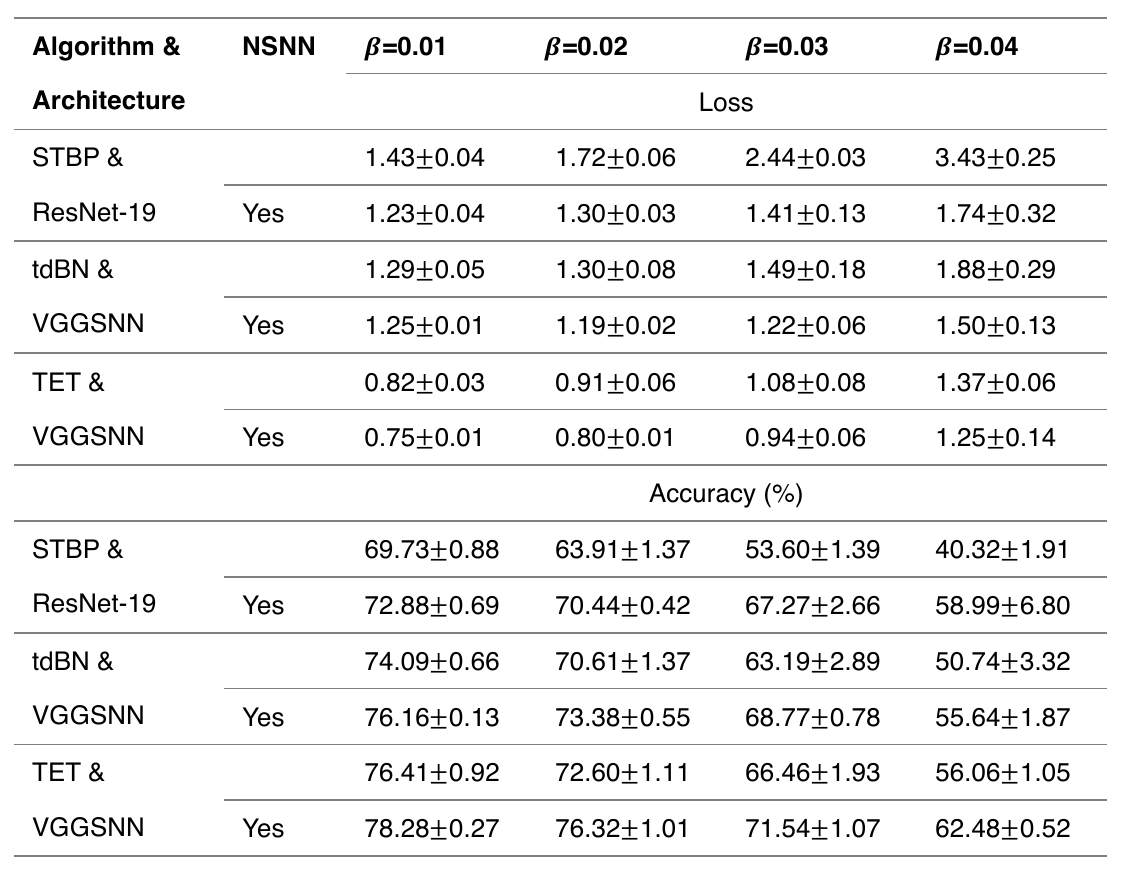}
    \label{tab:6}
\end{table}

\begin{table}[htb]
\centering
    \caption{
    	{\bf Hyper-parameter settings for recognition experiments.} 
    }
    \includegraphics[width=1 \textwidth]{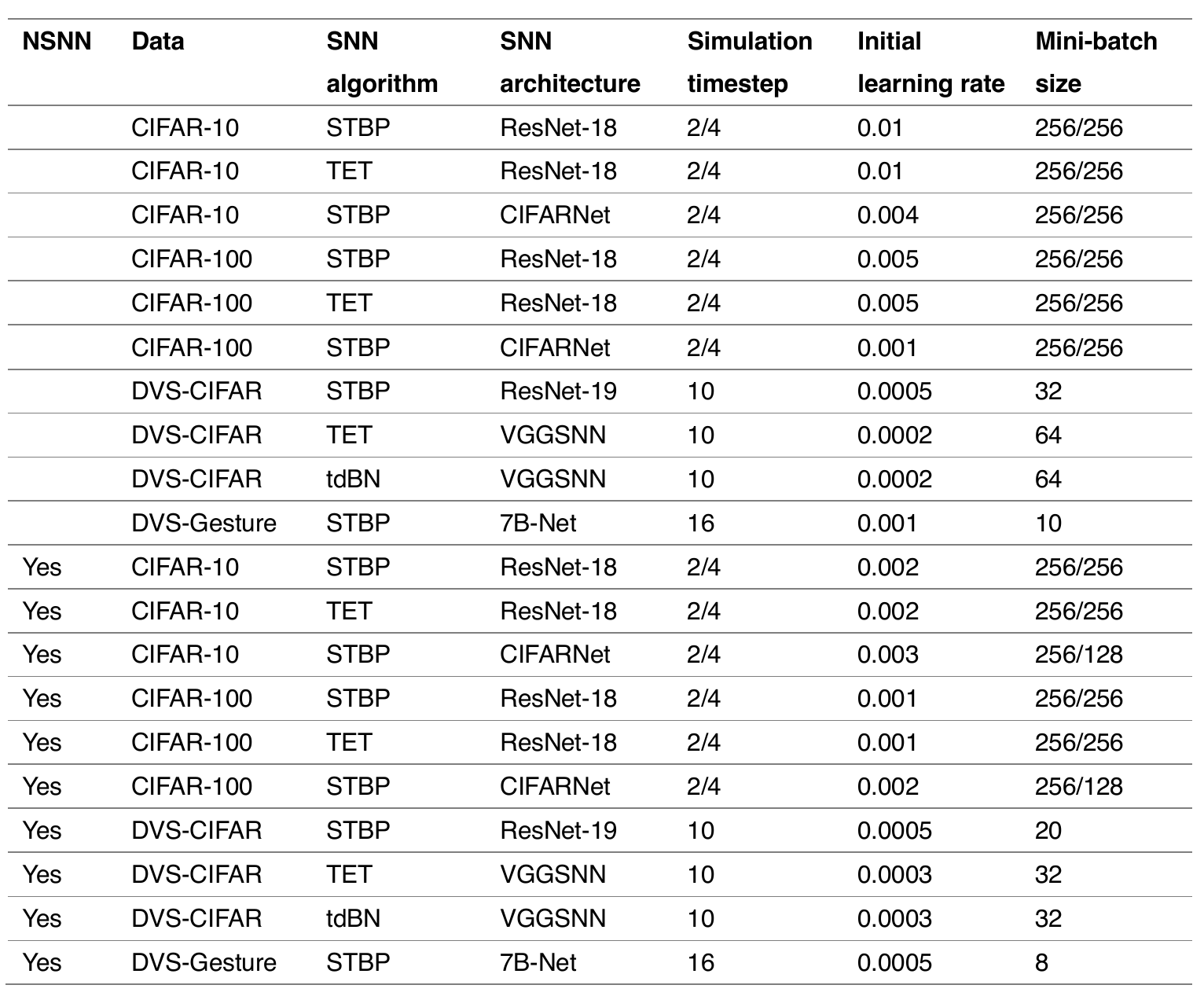}
    \label{tab:7}
\end{table}

\end{document}